\documentclass[journal]{IEEEtran}
\usepackage{times}
\usepackage{algorithm}
\usepackage{algpseudocode} 
\algrenewcommand\algorithmicindent{.9em}%
\usepackage{epsfig}
\usepackage{bm}
\usepackage{graphicx}
\usepackage{amsmath}
\usepackage{amssymb}
\usepackage{booktabs}
\usepackage{multirow}
\usepackage{enumitem}
\usepackage[pagebackref=false,breaklinks=true,letterpaper=true,colorlinks,urlcolor=blue,citecolor=blue,linkcolor=blue,bookmarks=false]{hyperref}

\usepackage{cite}

\ifCLASSINFOpdf
\else
\fi
\begin{document}
%
\title{Cascaded Regression Tracking: Towards Online Hard Distractor Discrimination}
%
%

\author{Ning Wang, 
	    Wengang Zhou, 
		Qi Tian,
		\emph{Fellow, IEEE},
	    and~Houqiang Li,
	    \emph{Senior Member, IEEE} 
\thanks{Ning Wang, Wengang Zhou, and Houqiang Li are with the CAS Key Laboratory of Technology in Geo-spatial Information Processing and Application System, Department of Electronic Engineering and Information Science, University of Science and Technology of China, Hefei, China. \protect\\ E-mail: wn6149@mail.ustc.edu.cn, \{zhwg, lihq\}@ustc.edu.cn.}
\thanks{Qi Tian is with the Huawei Noah's Ark Laboratory.\protect\\ E-mail: tian.qi1@huawei.com.}
\thanks{Corresponding authors: Wengang Zhou and Houqiang Li.}}

%
%

\markboth{IEEE TCSVT,~Vol.~XX, No.~XX, XX~2019}%
{Shell \MakeLowercase{\textit{et al.}}: Bare Demo of IEEEtran.cls for IEEE Journals}
%



\maketitle


\begin{abstract}
	
Visual tracking can be easily disturbed by similar surrounding objects. Such objects as hard distractors, even though being the minority among negative samples, increase the risk of target drift and model corruption, which deserve additional attention in online tracking and model update. 
To enhance the tracking robustness, in this paper, we propose a cascaded regression tracker with two sequential stages.
In the first stage, we filter out abundant easily-identified negative candidates via an efficient convolutional regression.
In the second stage, a discrete sampling based ridge regression is designed to double-check the remaining ambiguous hard samples, which serves as an alternative of fully-connected layers and benefits from the closed-form solver for efficient learning.
%
During the model update, we utilize the hard negative mining technique and an adaptive ridge regression scheme to improve the discrimination capability of the second-stage regressor.
Extensive experiments are conducted on 11 challenging tracking benchmarks including OTB-2013, OTB-2015, VOT2018, VOT2019, UAV123, Temple-Color, NfS, TrackingNet, LaSOT, UAV20L, and OxUvA.
The proposed method achieves state-of-the-art performance on prevalent benchmarks, while running in a real-time speed.

\end{abstract}

\begin{IEEEkeywords}
Visual tracking, regression tracking, cascaded framework, hard distractor.
\end{IEEEkeywords}

\IEEEpeerreviewmaketitle

\vspace{+0.00in}

\section{Introduction} \label{sec:intro}

\IEEEPARstart{A}{s} a fundamental task in computer vision, visual object tracking has received lots of attention over the last decades. 
It plays an important role in many applications such as autonomous driving, robotics, human-computer interaction, etc. 
In generic visual tracking, the target is arbitrary with only the initial bounding box available.
With such limited prior information, the tracker is still highly required to both model the target appearance and distinguish the negative samples on the fly, which is challenging due to the blurry boundary between appearance changes of the target itself and unforeseen similar distractors.

\begin{figure}[t]
	\centering
	\includegraphics[width=8.3cm]{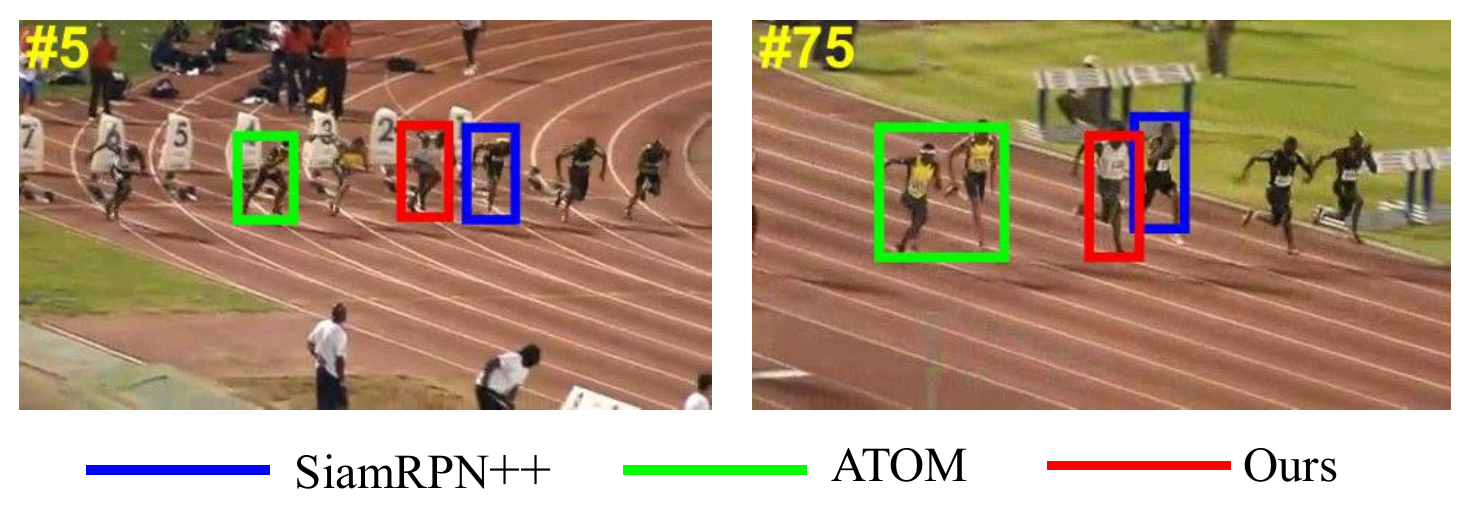}
	\includegraphics[width=8.8cm]{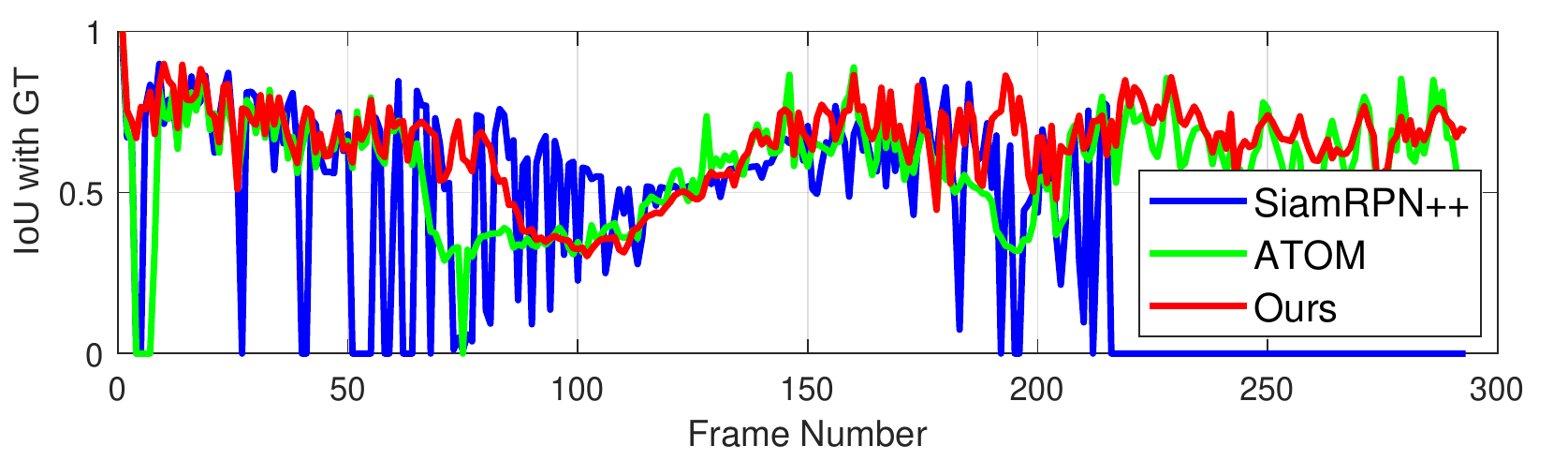}
	\caption{{\bf Top}: the tracking results of SiamRPN++ \cite{siamrpn++}, ATOM \cite{ATOM}, and our approach on the \emph{Bolt2} video. Note that the prior motion model is removed in these methods. {\bf Bottom}: the per-frame overlap between the tracking results and ground-truth bounding box on the \emph{Bolt2}. Without the cosine window, previous methods tend to switch between the target and distractors (e.g., $ 5 $-th and $ 75 $-th frame), while ours steadily tracks the target without drift.}
	\label{fig:1} 
\end{figure}

Recently, thanks to the strong representational power of deep CNN models, a simple two-stream template matching based Siamese pipeline \cite{SiamFC, SINT} has been proved effective in visual tracking, even without the online model update. 
However, as reported in the Visual Object Tracking (VOT) challenge \cite{VOT2018}, the robustness of Siamese trackers still has a margin with the discriminative trackers equipped with an update mechanism.
In the latest literature \cite{DiMP,SiameseUpdate}, substantial attentions have been cast to the updatable deep trackers with superior discrimination capability. 
Despite the rapid advances, in the tracking and updating stages, how to distinguish similar distractor objects from the target and effectively leverage these hard negative samples to boost the model discrimination capability still leaves exploration space. 
There exist vast uninformative samples that can be easily distinguished without much effort (i.e., easy sample), while a handful of distracting examples heavily mislead the tracker, enlarging the error accumulation and causing the tracking failures (Figure~\ref{fig:1}). 
These unexpectedly emerged distractors, even though being the minority, have a non-trivial effect on degrading the tracking performance, and deserve to be carefully checked \emph{online} for robust tracking.

In this paper, we propose a cascaded regression tracker, which consists of two sequential stages with different regression models for high-performance visual tracking. 
In the first stage, we employ an efficient convolutional regression \cite{ATOM} to densely predict all the searching locations, which filters out plentiful easy samples. 
In the second stage, we only consider the remaining ambiguous candidates and propose a discrete sampling based ridge regression for further discrimination.
The ridge regressor performs as an alternative of the fully-connected layer but exhibits superior efficiency thanks to its closed-form solution.
These two stages complement each other as follows. 
The dense prediction with the convolutional regressor in the first stage \cite{ATOM} covers a large search area, while its model tends to be disturbed by an overwhelming number of easy samples. 
In contrast, the second-stage regressor trained using the carefully selected hard samples naturally avoids the class-imbalance issue and yields better discrimination on distractors, while its sampling manner fails to perfectly cover the search area and will increase the computational cost when drawing plentiful candidates. 
By virtue of such a dense-to-discrete and coarse-to-fine two-tier verification, these two stages contribute to a superior robust tracking system. 
More importantly, both of them allow to update the corresponding models, achieving the \emph{online} adaptability.

During online tracking, to enhance the tracker discrimination, we employ the hard negative mining \cite{example-based_HNM,felzenszwalb2009object_HNM} for the second stage.
Moreover, we dynamically reweigh the training samples based on their reconstruction errors in an adaptive ridge regression formula, forcing the second-stage regressor to focus more on valuable samples. 
Benefited from the high robustness, the second-stage regressor is able to re-detect the lost target when the first stage fails to confidently track the target, and search a large region without excessively worrying about the risk of tracking drift.
As a consequence, our framework differs from most existing short-term trackers typically focusing on a limited search region with a prior cosine window to penalize the far-away distractors (e.g., Siamese trackers \cite{SiamFC,SiamRPN}). 
It is worth mentioning that our method shows outstanding performance on \emph{both} short-term and long-term tracking datasets without adding additional sophisticated modules thanks to our excellent online discrimination capability.

We summarize the contributions of our work as follows:
\begin{itemize}
	\setlength{\parskip}{0pt}	
	\item We propose a discrete sampling based ridge regression, which can flexibly absorb the online hard samples and is efficient to learn under a closed-form formula. Furthermore, we propose a cascaded regression tracker, which achieves favorable robustness via a dense-to-discrete large-scale search and a coarse-to-fine two-tire verification. 	
	\item To improve the online distractor discrimination, we propose an adaptive ridge regression to further exploit the valuable samples selected by the hard negative mining technique \cite{example-based_HNM,felzenszwalb2009object_HNM}. With the merit of promising discrimination, the second-stage regressor also serves as an effective re-detection module to complement the first stage.
	\item We extensively evaluate the proposed method on 11 short-term and long-term tracking benchmarks including OTB-2013 \cite{OTB-2013}, OTB-2015 \cite{OTB-2015}, Temple-Color \cite{TempleColor128}, UAV123 \cite{UAV123}, VOT2018 \cite{VOT2018}, VOT2019 \cite{VOT2019},  NfS \cite{NFSdataset}, TrackingNet \cite{2018trackingnet}, LaSOT \cite{LaSOT}, UAV20L \cite{UAV123}, and OxUvA \cite{2018longtermBenchmark}. The proposed approach exhibits state-of-the-art performance on prevalent datasets with a real-time speed. 
\end{itemize}

In the following of the paper, we first survey related works in Section \ref{sec:related}. Then, we elaborate the proposed cascaded framework in Section \ref{sec:method}. After that, we evaluate our method with extensive experiments in Section \ref{sec:experiments}. Finally, we conclude this work in Section \ref{sec:conclusion}.

\section{Related Work} \label{sec:related}

In recent years, the Siamese network has gained significant popularity in visual tracking, which deals with the tracking task by searching for the image region most similar to the initial template \cite{SiamFC,SINT}. 
The GOTURN algorithm \cite{GOTURN} adopts a Siamese pipeline to regress the target bounding box.
By introducing the RPN module \cite{SiamRPN,DaSiamRPN}, ensemble learning \cite{SASiam}, attention mechanism \cite{RASNet}, and target-aware formulation \cite{TADT}, the Siamese trackers gain substantial improvements.
Besides visual tracking, similar ideas such as one-shot learning \cite{caelles2017one-shotSegmentation} and online adaptation scheme \cite{voigtlaender2017onlineSegmentation} are widely explored in the video object segmentation task.
Without video annotations, the unsupervised deep tracking framework is explored in UDT \cite{UDT}.
In \cite{siamrpn++}, SiamRPN++ adopts a deeper backbone network to achieve superior performance.
By switching multiple Siamese trackers using an agent network, POST tracker \cite{POST} achieves a good balance of accuracy and efficiency.
%
%
Recently, model update mechanisms are incorporated with the Siamese network \cite{GCT,Dsiam,MemTrack,SiameseUpdate}. 
However, these approaches mainly focus on the template adaptation and still fail to exploit the background context. 
Since most Siamese trackers ignore the informative negative samples for discrimination enhancement, they tend to drift when similar distractors appear. 
Recently, the cascaded framework has been investigated within the Siamese tracking framework \cite{CRPN,SPM}. 
SPM \cite{SPM} combines the SiamRPN with a relation network to further classify the candidates. C-RPN \cite{CRPN} utilizes cascaded region proposal networks for accurate target localization. 
Nevertheless, they do not involve the online model update. The overlook of online emerged samples heavily limits the performance.
In other words, how to take advantage of the hard negative samples to distinguish potential distractors in future frames is ignored in the recent cascaded frameworks. 
Compared with them, the main distinction of this work is that our cascaded framework is built on two complementary regression models, both of which are able to absorb the online samples for the persistent model update.

\begin{figure*}[t]
	\centering
	\includegraphics[width=18.1cm]{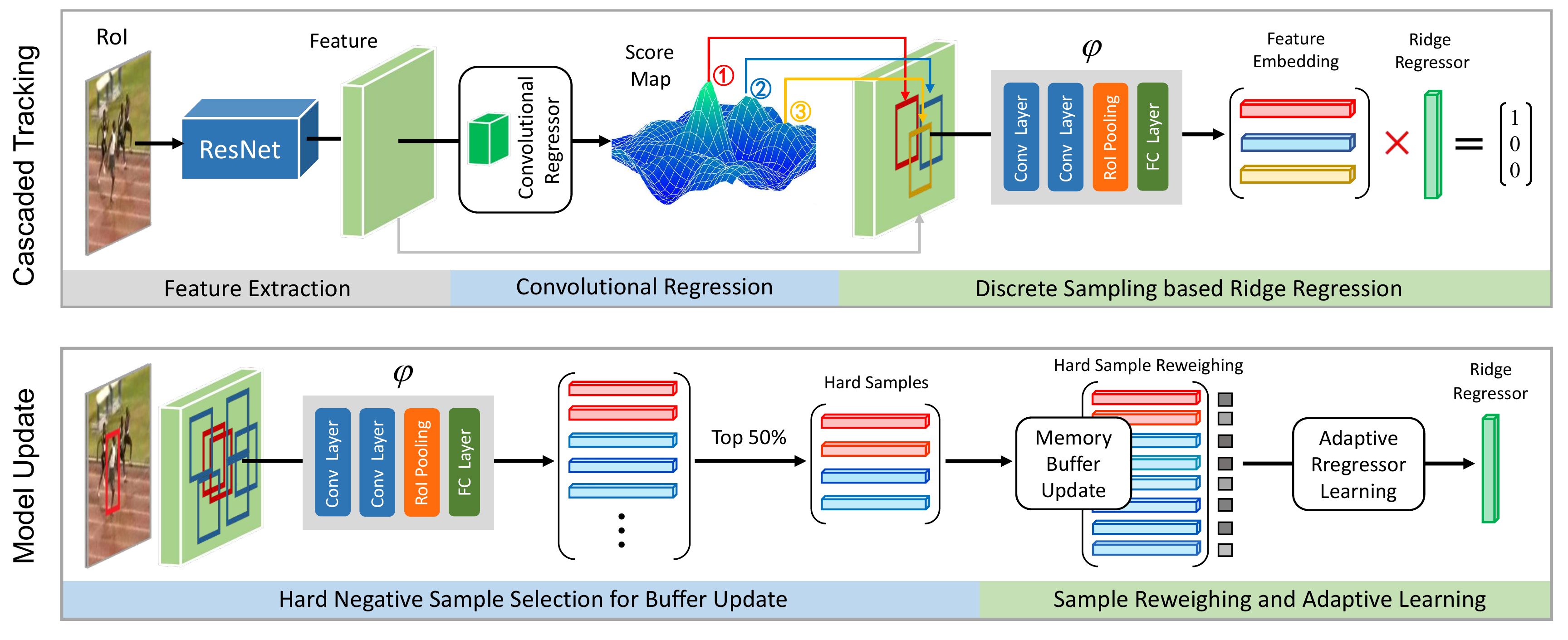}
	\caption{{\bf Top}: an overview of our cascaded regression tracker. In the first stage, we employ a convolutional regression for dense response prediction. In the second stage, a discrete sampling based ridge regression is designed to discriminate the ambiguous candidates. {\bf Bottom}: online model update for the second-stage regressor. Based on the tracking result, only hard negative samples are selected to update the memory buffer. Besides, the training samples in the buffer are dynamically reweighed for adaptive regressor learning.}
	\label{fig:main} 
\end{figure*}

Another popular tracking family is the regression based approach, which generally regresses a large Region of Interest (RoI) to a response map for target localization.
The Correlation Filter (CF) solves the ridge regression in the Fourier domain, showing extremely attractive efficiency \cite{KCF,DSST,HCF,CSR-DCF,target_response_CF,SCT,ACFN,Context-AwareCorrelationFilter,MCCT,C-COT}. 
To alleviate the unwanted boundary effect, regularization terms \cite{ASRCF,SRDCF,STRCF} and background-aware formulation \cite{BACF} are proposed. 
ECO tracker \cite{ECO} introduces a factorized convolution operator, a generative sample space model, and the sparse update strategy to further boost the efficiency of correlation tracking.
Recently, by jointly compressing and transferring the heavyweight feature extractors in deep CF trackers, CPU real-time efficiency is also feasible \cite{CF-VGG}. 
Besides CF, with the recent astonishing development of deep learning, convolutional regression gains an increasing attention in visual tracking \cite{CREST,DSLT,ATOM,DiMP}. 
In these approaches, a CNN kernel is learned to convolve with the RoI feature for response generation, which effectively avoids the boundary effect in CF.
Unfortunately, this convolutional formulation does not have a closed-form solution, and needs the gradient back-propagation to learn the filter.
Besides, the large RoI size in the regression approach brings in the class-imbalance issue.
In CREST \cite{CREST}, residual terms are incorporated into the convolutional regression to cope with the target appearance changes.
DSLT \cite{DSLT} introduces shrinkage loss to balance the training samples in the convolutional regression.
To accelerate the kernel learning process, ATOM \cite{ATOM} exploits the conjugate gradient in the deep learning framework. 
The recent DiMP approach \cite{DiMP} proposes an iteratively optimized discriminative model for classification and trains the whole framework in an end-to-end manner.
Despite the recent progress, the discrimination capability in regression trackers, especially for hard distractors, still leaves room for improvement.

In contrast to the aforementioned regression methods that generate a dense prediction, previous discriminative trackers learn a binary classifier to classify the discretely sampled candidates for tracking (e.g., MDNet \cite{MDNet}).
In spite of their shallow backbone networks and limited discrete samples (e.g., 256 candidates per frame), by an effective model update with hard negative mining, these approaches \cite{MDNet,RTMDNet,VITAL} still exhibit impressive robustness on various tracking benchmarks, suggesting the importance of online learning.

Our proposed approach is partially inspired by the above observations to retain both the dense and discrete predictions in a coarse-to-fine manner. 
Hard negative mining, as a powerful technique in object detection \cite{example-based_HNM,felzenszwalb2009object_HNM}, has been successfully equipped into some discrete sampling based visual trackers such as MDNet \cite{MDNet}. 
However, existing regression based trackers fail to effectively explore the hard negative samples since they train the regression model using densely sampled candidates and generally equally weigh them.
The recent ATOM tracker \cite{ATOM} reduces the training weights of easy samples to focus on the valuable negative samples to some extent, but we observe that it still struggles to distinguish hard distractors.
In this work, our first stage densely searches a large RoI to generate high-quality proposals, while the second stage is more flexible in the model update and hard negative mining to better distinguish the hard negative samples.
Even though aiming at predicting discrete samples, unlike \cite{MDNet,RTMDNet,VITAL} that leverage fully-connected layers for classification, we learn an efficient closed-form solver in the feed-forward pass without back-propagation, potentially alleviating the overfitting issue due to much fewer parameters to be optimized online.
By design, we absorb the strength of both regression trackers and discrete sampling based tracking-by-detection approaches to form a unified cascaded tracking framework.
Our method is also motivated by the two-stage framework in object detection (e.g., faster RCNN \cite{FasterRCNN}), which has witnessed tremendous success in recent years. 
Differently, we exploit two regression models specially designed for the online tracking task with an incremental model update.

\section{Methodology} \label{sec:method}

In Figure~\ref{fig:main} (top), we show an overview of the proposed cascaded tracker. In the first stage, a convolutional regressor densely predicts the target location over a large RoI. 
Then, the ambiguous proposals are fed to the second regression stage for further discrimination. 
Under such a dense-to-discrete and coarse-to-fine verification, the proposed tracking framework achieves favorable tracking robustness.
In Figure~\ref{fig:main} (bottom), we exhibit the online update process of the second-stage regression model.
By virtue of the hard negative mining and an adaptive ridge regression formulation, the learned regressor is readily ready for distinguishing hard distractors.

In the following, we first review the regression based tracking in Section~\ref{review regression tracking} for the sake of completeness. 
In Section~\ref{discrete ridge regression}, we present our discrete sampling based ridge regression and provide a detailed analysis in comparison with the previous methods. 
Then, in Section~\ref{online tracking}, we depict the cascaded regression tracking and re-detection mechanism.
Finally, we introduce the details of the online model update in Section~\ref{model update}.

\subsection{Revisiting Regression Tracking} \label{review regression tracking}
In this subsection, we briefly review the correlation filter and convolutional regression.

{\flushleft \bf Correlation Filter.} The correlation filter (CF) \cite{DSST,KCF} tackles visual tracking by solving the following regression problem:
\begin{equation}\label{eq1}
\min_{{\bf W}_{\text{CF}}}\|{\bf X}\star{{\bf W}_{\text{CF}}}-{{\bf Y}_{\text{G}}}\|^{2}_{2} + \lambda\|{{\bf W}_{\text{CF}}}\|^{2}_{2},
\end{equation}
where $ \star $ denotes the circular correlation, $ \lambda $ is a regularization parameter that controls overfitting, $ {\bf X}\in\mathbb{R}^{M\times N\times C} $ is the feature map of the RoI patch, $ {\bf Y}_{\text{G}}\in\mathbb{R}^{M\times N} $ is the Gaussian-shaped label, and $ {{\bf W}_{\text{CF}}}\in\mathbb{R}^{M\times N\times C} $ is the desired correlation filter.

Let $ \bf A $ denote the data matrix that contains all the circulant shifts of the base feature representation $ \bf X $. Then, the circular correlation $ {\bf X}\star{{\bf W}_{\text{CF}}} $ is equal to $ {\bf A}{{\bf W}_{\text{CF}}} $, and the filter $ {\bf W}_{\text{CF}} $ has the following closed-form solution \cite{rifkin2003regularized,KCF,bertinetto2018meta}: 
\begin{equation}\label{eq2}
{{\bf W}_{\text{CF}}} = ({\bf A}^{\mathrm{T}}{\bf A}+\lambda {\bf I})^{-1}{\bf A}^{\mathrm{T}}{{\bf Y}_{\text{G}}},
\end{equation}
where $ \bf I $ is the identity matrix. Due to the circulant structure of $ \bf A $, it can be diagonalized via $ {\bf A}={\bf F}~\text{diag}(\hat{\bf X})~{\bf F}^{\mathrm{H}} $, where $ \hat{\bf X} $ is the Discrete Fourier Transform (DFT) of $ \bf X $, $\bf F $ is the DFT matrix and $ {\bf F}^{\mathrm{H}} $ is the Hermitian transpose of $ \bf F $. 
Therefore, Eq.~\ref{eq2} results in a very efficient element-wise multiplication solution in the Fourier domain without matrix inversion.
Please refer to \cite{KCF} for more details.

{\flushleft \bf Convolutional Regression.} The convolutional regression \cite{CREST,DSLT,ATOM} considers the following minimization problem:
\begin{equation}\label{eq3}
\min_{{\bf W}_{\text{Conv}}} \|{\bf X}\ast {{\bf W}_{\text{Conv}}} - {{\bf Y}_{\text{G}}}\|^{2}_{2} + \lambda\|{{\bf W}_{\text{Conv}}}\|^{2}_{2}.
\end{equation}
Different from the circular correlation in Eq.~\ref{eq1}, the $ \ast $ operation in Eq.~\ref{eq3} denotes the standard multi-channel convolution, which is the core component in CNNs.

Without a closed-form formula, the solution of Eq.~\ref{eq3} can be optimized via the standard gradient descent as follows:
\begin{equation}\label{solution of eq3}
{\bf W}_{\text{Conv}}^{i+1} = {\bf W}_{\text{Conv}}^{i} - \alpha \nabla {\cal L}({\bf W}_{\text{Conv}}^{i}),
\end{equation}
where $ \alpha $ is the learning rate of the gradient descent and $ {\cal L}(\cdot) $ denotes regression error presented in Eq.~\ref{eq3}.
Given the feature map $ {\bf X}\in\mathbb{R}^{M\times N\times C} $, the learned filter (or convolutional kernel) $ {{\bf W}_{\text{Conv}}}\in\mathbb{R}^{m\times n\times C} $ regresses the feature map $ \bf X $ to the desired Gaussian label $ {\bf Y}_{\text{G}} $. 
Note that the correlation filter $ {{\bf W}_{\text{CF}}} $ in Eq.~\ref{eq1} has the same spatial size with $ \bf X $, while the convolutional filter $ {{\bf W}_{\text{Conv}}} $ requires to be smaller than $ \bf X $, i.e., $ m<M,~n<N $, as shown in Figure~\ref{fig:comparison} (b).

\subsection{Discrete Sampling based Ridge Regression} \label{discrete ridge regression}
\begin{figure}[t]
	\centering
	\includegraphics[width=8.7cm]{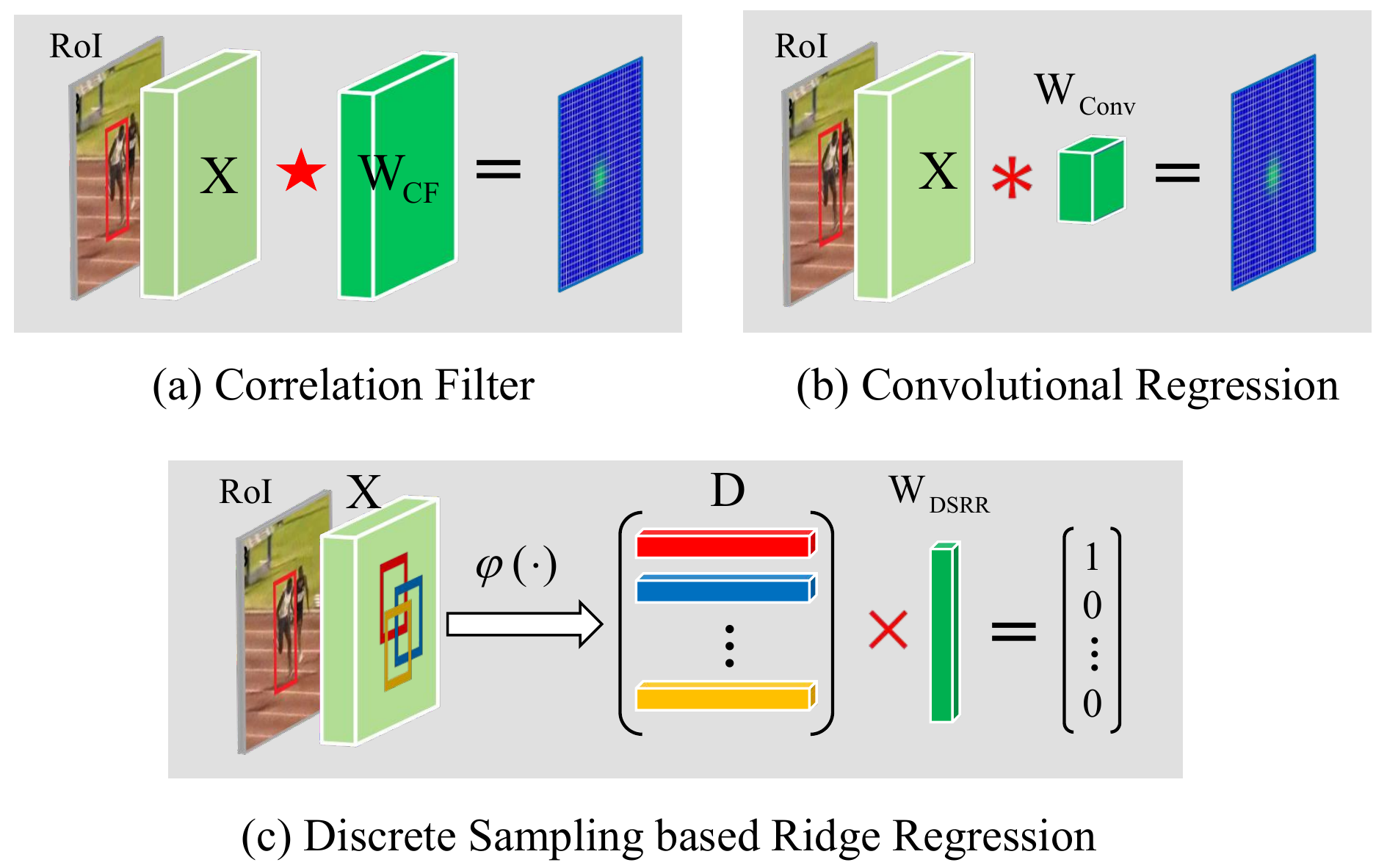}
	\caption{Illustration of correlation filter, convolutional regression, and our discrete sampling based ridge regression.}
	\label{fig:comparison} 
\end{figure}

In the CF and convolutional regression, the learned filters regress the RoI to a dense response map.
This continuous prediction generally brings in the class-imbalance issue \cite{DSLT}, where plentiful uninformative samples will overwhelm the valuable ones in the filter training.
Actually, there is no need to limit ourselves to the dense prediction in a regression scheme.
To focus on the hard samples, we propose a simple, flexible yet effective Discrete Sampling based Ridge Regression (DSRR). 
The \emph{discrete} lies in two aspects: (1) The training data are sampled discretely (Figure~\ref{fig:comparison} (c)), which is similar to the classic classification based tracking approach \cite{MDNet}. 
By carefully selecting the training samples, the learned filter pays more attention to the hard negative samples and naturally avoids the class-imbalance issue. 
(2) The label is discrete (binary) instead of the soft Gaussian shape, which introduces the label margin between positive and hard negative samples.
As shown in Figure~\ref{fig:comparison} (c), the learned discrete ridge regressor can be interpreted as a fully-connected layer with a single node, but provides a fast solution in a single pass to learn the model instead of learning with time-consuming back-propagation.

To train this regressor, we represent each sample by a high-dimensional feature embedding via a CNN mapping function $ \varphi({\bf X}, {\bf B}_i) $, whose inputs consist of the base feature map $ \bf X $ and the $ i $-th sample's bounding box $ {\bf B}_i\in\mathbb{R}^{4} $. 
These training samples are discretely sampled with binary labels, representing the target or background. 
As shown in Figure~\ref{fig:main}, the mapping function $ \varphi(\cdot) $ first refines the backbone feature $ \bf X $ through two convolutional layers, and further generates the feature embedding via an RoI pooling operation followed by a fully-connected layer.
Then we assemble these feature embeddings to form the data matrix $ {\bf D}=[\varphi({\bf X},{\bf B}_1), \cdots,\varphi({\bf X},{\bf B}_P)]^{\mathrm{T}} \in\mathbb{R}^{P\times L} $, which contains $ P $ embeddings and each of them is $ L$-dimensional. 
Based on the overlap ratios between candidates' boxes $ {\bf B} $ and ground-truth box, these feature embeddings are assigned by positive or negative labels. Leveraging data matrix $ \bf D $ and its binary label $ {\bf Y}_{\text{B}} $, the discrete sampling based ridge regressor $ {\bf W}_{\text{DSRR}} $ can be obtained by solving the following minimization problem:
\begin{equation}\label{eq4}
\min_{{\bf W}_{\text{DSRR}}}\|{\bf D} {{\bf W}_{\text{DSRR}}} -{{\bf Y}_{\text{B}}}\|^{2}_{2} + \lambda\|{{\bf W}_{\text{DSRR}}}\|^{2}_{2},
\end{equation}
where $ {\bf Y}_{\text{B}}\in\mathbb{R}^{P} $ is the binary label.

{\flushleft \bf Primal Domain.} Since Eq.~\ref{eq4} still follows the standard ridge regression, similar to Eq.~\ref{eq2}, it has the closed-form solution $ {\bf W}_\text{DSRR} = ({\bf D}^{\mathrm{T}}{\bf D}+\lambda {\bf I})^{-1}{\bf D}^{\mathrm{T}}{{\bf Y}_{\text{B}}} $. 
Compared with the solution to CF, the main advantage is that this data matrix $ \bf D $ no longer contains fake (cyclically shifted) samples, while the tradeoff is that the Fourier domain solution becomes unfeasible. 
In the above solution, the main computational burdern lies in the matrix inverse, whose time complexity is $ O(L^3) $ for the matrix $ ({\bf D}^{\mathrm{T}}{\bf D}+\lambda {\bf I})\in \mathbb{R}^{L\times L}$.

{\flushleft \bf Dual Domain.}
Eq.~\ref{eq4} can also be solved in the dual domain, where the regressor $ {\bf W}_\text{DSRR} $ is expressed by a linear combination of the samples, i.e., $ {\bf W}_\text{DSRR} ={\bf D}^{\mathrm{T}}\bm{\alpha}$.
The variables under optimization are thus $ \bm{\alpha} $ instead of $ {\bf W}_{\text{DSRR}} $.
The dual variables $ \bm{\alpha} $ can be solved by $ \bm{\alpha} = ({\bf D}{\bf D}^{\mathrm{T}}+\lambda {\bf I})^{-1}{{\bf Y}_{\text{B}}} $ \cite{rifkin2003regularized}.
Therefore, the ridge regressor can be computed in the dual domain as follows:
\begin{equation}\label{Eq5}
{\bf W}_\text{DSRR} = {\bf D}^{\mathrm{T}}\bm{\alpha} = {\bf D}^{\mathrm{T}}({\bf D}{\bf D}^{\mathrm{T}}+\lambda {\bf I})^{-1}{{\bf Y}_{\text{B}}}.
\end{equation}  
Since $ ({\bf D}{\bf D}^{\mathrm{T}} + \lambda {\bf I})\in \mathbb{R}^{P\times P}$, the matrix inverse in Eq.~\ref{Eq5} has the time complexity of $ O(P^3) $ instead of $ O(L^3) $ in primal domain, which relates to the sample number $ P $ instead of feature dimension $ L $. Thanks to the limited number of hard examples, a small $ K $ is generally practicable. 
While in case of a low feature dimension $ L $, the primal domain solution will be more efficient.
Overall, depending on the sizes of $ L $ and $ P $, we can always find a good efficiency balance between the primal and dual solutions.

{\flushleft \bf Offline Training.} In the training stage, we aim to learn a CNN function $ \varphi(\cdot) $ to ensure the learned feature representation suitable for the designed ridge regression. 
To this end, we adopt a Siamese-like pipeline in the training stage, where the template branch is utilized to learn the ridge regressor while the search branch is used to generate plentiful test candidates for loss computation. 
In the large Region of Interest (RoI), we randomly draw plentiful samples.
The positive and negative samples are collected following the ratio of 1~$:$~3, which have $ \geqslant$~0.7 and $ \leqslant $~0.5 overlap ratios with ground-truth bounding boxes, respectively.
In our experiment, the total sample number is 400 in each frame, i.e., 100 positive samples and 300 negative samples.

Instead of using the prototype $ \varphi(\cdot) $ in Figure~\ref{fig:main} for simplicity, to achieve better performance, we exploit the multi-scale feature representations from both $ \mathrm{Block3} $ and $ \mathrm{Block4} $ of the ResNet-18 \cite{ResNet} as the inputs of two individual $ \varphi(\cdot) $ networks.
The Precise RoI Pooling ($ \mathrm{PrPool} $) \cite{IoUNet} is utilized in $ \varphi(\cdot) $ to crop the $ \mathrm{Block3} $ and $ \mathrm{Block4} $ features, whose output sizes are 5$\times$5 and 3$\times$3, respectively. 
The following fully-connected layer maps the pooled features to a 256-dimensional feature vector.
Finally, the $ \mathrm{Block3} $ and $ \mathrm{Block4} $ feature vectors are concatenated along the channel dimension as the 512-dimensional output feature embedding.

Thanks to the closed-form solution of ridge regression, it can be embedded as a differentiable layer for end-to-end training. 
Leveraging the regressor $ {\bf W}_\text{DSRR} $ learned via template branch, the regression scores of the test candidates in the search branch can be calculated by $ {\bf Y}_\text{Test} = {\bf D}_\text{Test}{\bf W}_\text{DSRR} $.
To train the network $ \varphi(\cdot) $, we adopt the standard $ L_2 $ loss as the training objective: $ \ell = \|{\bf Y}_\text{Test}- {\bf Y}_\text{GT}\|^{2}_{2} $, where $ {\bf Y}_\text{GT} $ is the ground-truth binary label of the test samples.
After offline training, $ \varphi(\cdot) $ is fixed in the tracking stage.

\begin{figure}[t]
	\centering
	\includegraphics[width=7.3cm]{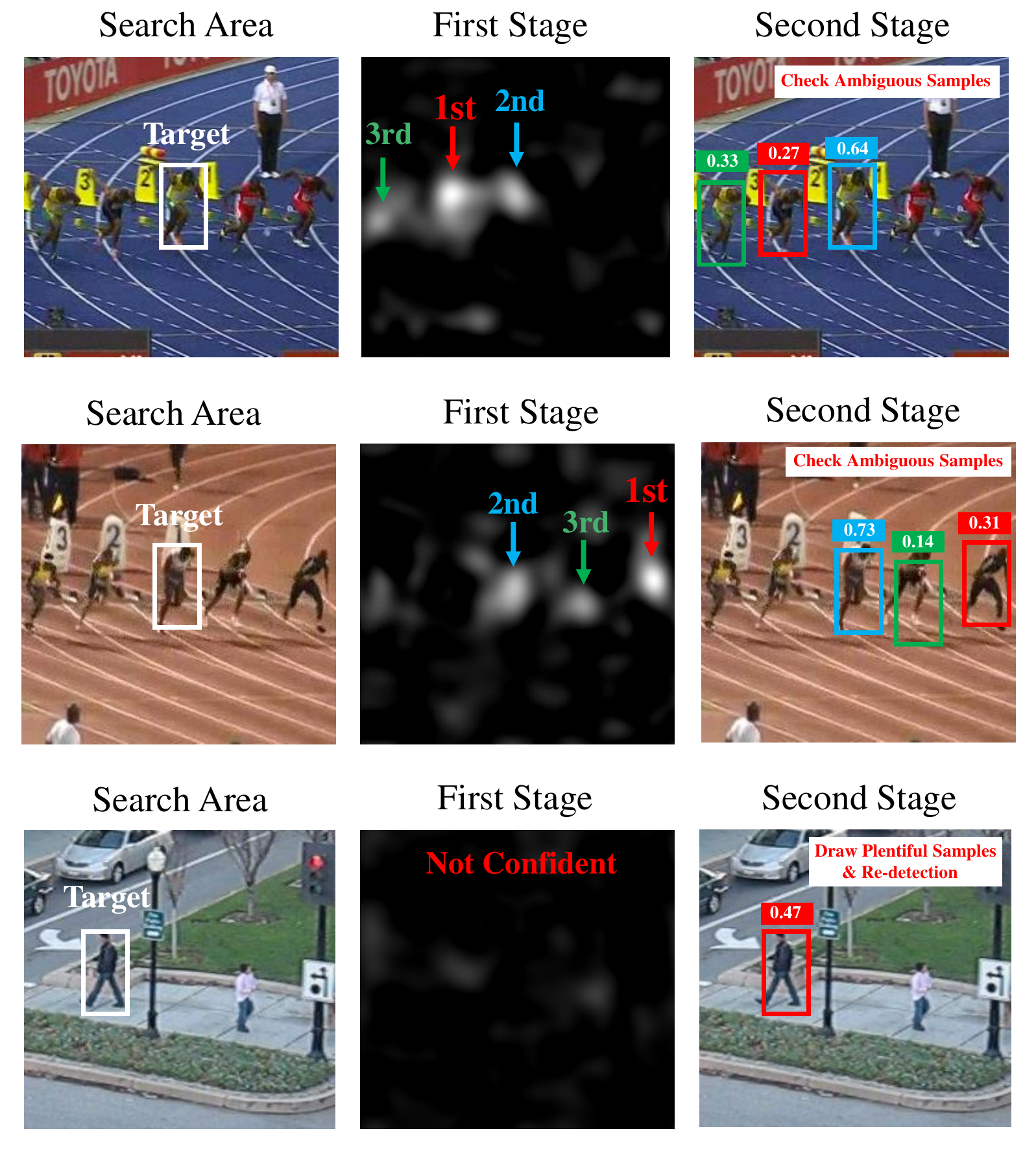}
	\caption{Tracking examples of the proposed cascaded framework. In the first stage, we select top three peaks as the high-quality proposals, which are further checked via the second stage. If the first stage fails to predict confidently, we draw plentiful candidates in the second stage for target re-detection.}
	\label{fig:tracking instance} 
\end{figure}

{\flushleft \bf Connection with Related Methods.}
We compare the CF, convolutional regression, and our discrete sampling based ridge regression in the following 4 aspects.
(1) {\bf Efficiency.} Convolutional regression typically requires gradient back-propagation to learn the filter.
CF exploits the closed-form solution in the Fourier domain, showing extremely attractive efficiency. 
The proposed DSRR also has a closed-form solution, yielding satisfactory efficiency. 
(2) {\bf Label.} Both CF and convolutional regression predict dense response scores. In contrast, our approach considers discrete proposals, which is flexible to focus on the hard examples and eliminate the class-imbalance issue. 
(3) {\bf Effectiveness.} The performance of CF is heavily limited by the boundary effect, i.e., the data matrix $\bf  A $ consists of plentiful fake samples. In contrast, the convolutional regression and our DSRR are learned using \emph{real} samples. 
(4) {\bf Flexibility.} The CF can only detect the RoI with a fixed size (Figure~\ref{fig:comparison}). In contrast, the convolutional regression and DSRR are more flexible, which can be applied to the RoI of any size and explore a larger area when necessary (e.g., target out-of-view). 
Considering the above characteristics, we choose the convolutional regression and discrete ridge regression as the first and second stages in our approach, respectively.

Our discrete ridge regression also shares partial similarity with the classification based approach (MDNet \cite{RTMDNet}). The main distinction is that we learn a closed-form solver to \emph{regress} the proposals instead of leveraging several fully-connected (FC) layers to \emph{classify} them, which is much more efficient via a feed-forward computation without back-propagation to update the FC parameters.

\subsection{Online Tracking} \label{online tracking}
{\flushleft \bf Cascaded Regression Tracking.}
Before tracking, we first learn the aforementioned two regressors using the initial frame. 
For the first stage, instead of adopting the stochastic gradient descent (SGD) to learn the convolutional filter, we follow Danelljan \emph{et al.} \cite{ATOM} to use Newton-Gaussian descent as the optimization strategy for fast convergence, and learn a 4$ \times$4 kernel to regress the Gaussian response map.
To learn the second-stage regressor, based on the initial ground-truth label, we crop the positive and negative samples following a ratio of 1~$:$~3 to form the data matrix.
Then, the discrete ridge regressor $ {\bf W}_\text{DSRR} $ is obtained by the primal or dual solution, depending on the sample number and feature dimension. In the initial few frames, the sample number is smaller than the feature dimension (i.e., $P < L$), and we choose the dual domain. With the arrival of new frames, if $ P > L$, we switch to the primal domain.

\begin{figure}[t]
	\centering
	\includegraphics[width=8.6cm]{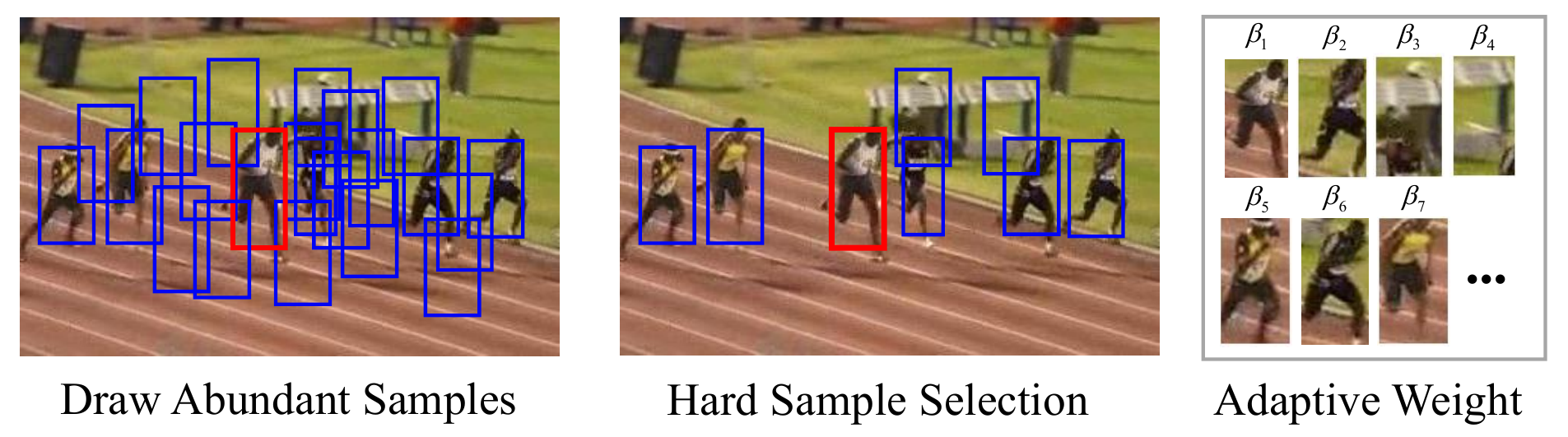}
	\caption{An illustration of the hard negative mining and sample reweigh for the proposed  adaptive ridge regression learning.}
	\label{fig:hard_negative} 
\end{figure}

During online tracking, in each frame, top-3 peaks in the first stage's score map are selected, and the corresponding proposals are fed to the network $ \varphi(\cdot) $ to generate the feature embeddings.
These ambiguous proposals are further checked by the second-stage regressor, as shown in Figure \ref{fig:tracking instance}.
Finally, we equally combine the prediction scores of these two stages, and select the highest proposal as the current target. 
After target localization, we utilize the IoU predictor proposed in \cite{ATOM} to further refine the target scale.

{\flushleft \bf Cascaded Re-detection.}
As a common strategy in many visual trackers \cite{DiMP,ATOM,MDNet}, we set two reliability thresholds $ \tau_1 $ and $ \tau_2 $ for the two regressors, respectively.
In case that the first stage cannot confidently predict the target, i.e., the highest response score is lower than $ \tau_1 $, we sample abundant candidates (512 per frame) and leverage the second-stage regressor for re-detection, as shown in Figure \ref{fig:tracking instance}. 
If the confidence score of the re-detected target exceeds $ \tau_2 $, we regard it as the target. Otherwise, we keep the target position as in the previous frame.  
Since the backbone features are shared, this re-detection process and the following model update only involve a slight computational burden.

\subsection{Online Model Update} \label{model update}
{\flushleft \bf Hard Negative Mining.}
Model update is the core component for discriminating the online distractors.
To alleviate the corruption of the memory buffer, we only collect the training samples in reliable frames.
Here, a \emph{reliable} frame represents that both two regressors predict confidently, i.e., their estimated scores exceed $ \tau_1 $ and $ \tau_2 $, respectively.

The first stage is incrementally updated by the Gauss-Newton descent using newly collected RoI samples following \cite{ATOM}. The second stage is expected to distinguish ambiguous samples.
To this end, we discretely draw two times of the desired negative samples and select only half of them with a high regression score, as shown in Figure~\ref{fig:hard_negative}.
These hard training samples are added to the buffer for the model update.

\begin{algorithm}[t] 
	\caption{Online Model Update}
	\begin{algorithmic}[1] 
		\State \textbf{Input}: Video sequence and initial ground-truth.
		\State Initialize two regressors $ {\bf W}_\text{Conv} $ and $ {\bf W}_\text{DSRR} $.
		\State Initialize the reliable frame buffer $ {\cal T}=\{1\} $.
		\For{$t=2 $ \textbf{to} $N$}                          
		\State  Conduct cascaded regression tracking;   \Comment{{\small Section 3.3}}
		
		\If{current result is reliable} 
		\State  Collect the current RoI region $ {\cal R}_{t} $;  \Comment{{\small first stage}}
		\State  Draw pos/neg samples $ {\cal S}^{+}_{t} $ and $ {\cal S}^{-}_{t} $; \Comment{{\small second stage}}
		\State  Drop $ 50\% $ easy negative samples from $ {\cal S}^{-}_{t} $;
		\State  $ {\cal T}\leftarrow {\cal T}\cup\{t\} $; \Comment{ {\small merge the reliable frame} }
		
		\If{$ |{\cal T}|> \gamma $}   \Comment{ {\small maintain a fixed buffer size } }
		\State  $ {\cal T}\leftarrow {\cal T}\setminus\{\text{min}_{k\in{\cal T}}k\} $; \Comment{{\small drop the oldest index}}
		\EndIf
		
		\EndIf
		
		\If{$ t $ \text{mod} 10 $==$ 0 } \Comment{ {\small sparse model update}}
		\State  Update $ {\bf W}_\text{Conv} $ using $ {\cal R}_{k\in{\cal T}} $; \Comment{{\small first stage}}
		\State 	Update $ {\bf W}_\text{DSRR} $ using $ {\cal S}^{+/-}_{k\in{\cal T}} $;  \Comment{{\small second stage}}
		\EndIf
		\EndFor
	\end{algorithmic}
	\label{code1}
\end{algorithm}

{\flushleft \bf Adaptive Ridge Regression.} 
For the second stage, under consistent model update, the ambiguity degrees of different training samples dynamically change.  
Therefore, we further assign a weight $ \beta_i $ to each training sample $ \varphi({\bf X},{\bf B}_i) $ in the memory buffer. 
As a result, the discrete ridge regression is re-formulated as follows:
\begin{equation}\label{eq6}
\min_{{\bf W}_{\text{DSRR}}} \sum_i \beta_i\|{\varphi({{\bf X}, {\bf B}_i})} {{\bf W}_{\text{DSRR}}} -{y_i}\|^{2}_{2} + \lambda\|{{\bf W}_{\text{DSRR}}}\|^{2}_{2}.
\end{equation}
By defining a weight matrix $ {\bf M} = [\sqrt{\beta_1},\cdots,\sqrt{\beta_P}]^{\mathrm{T}} $, Eq.~\ref{eq6} can be converted into the matrix form as follows: 
\begin{equation}\label{eq6_new}
\min_{{\bf W}_{\text{DSRR}}} \| {\bf M} \odot {\bf D} {{\bf W}_{\text{DSRR}}} -{\bf M} \odot {\bf Y}_{\text{B}}\|^{2}_{2} + \lambda\|{{\bf W}_{\text{DSRR}}}\|^{2}_{2}.
\end{equation}
As a result, the the solution to Eq.~\ref{eq6} can be computed by
\begin{equation}\label{eq7}
{\bf W}_\text{DSRR} = (\widetilde{\bf D}^{\mathrm{T}}\widetilde{\bf D}+\lambda {\bf I})^{-1}\widetilde{\bf D}^{\mathrm{T}}{\widetilde{{\bf Y}_{\text{B}}}} =  \widetilde{\bf D}^{\mathrm{T}}(\widetilde{\bf D}\widetilde{\bf D}^{\mathrm{T}}+\lambda I)^{-1}{\widetilde{{\bf Y}_\text{B}}},
\end{equation}
where $ \widetilde{\bf D} = {\bf M}\odot {\bf D} $, $ \widetilde{{\bf Y}_{\text{B}}} = {\bf M}\odot {\bf Y}_\text{B} $, and $ \odot $ is the element-wise product.
We empirically define the weight matrix $ \bf M $ as the reconstruction error of the sample label by previous ridge regressor, as follows:
\begin{equation}\label{eq10}
{\bf M} = \text{norm}\left({\bf Y}_\text{B} - {\bf D}{\bf W}^{t-1}_\text{DSRR}\right) \cdot P,
\end{equation}
where $ {\bf W}^{t-1}_\text{DSRR} $ is the ridge regressor in the previous frame, $ \text{norm}({\bf x}) = {\bf x} / \|{\bf x}\|_1 $ denotes $L_1$ normlization, and $ P $ is the total sample number in the data matrix. 
Intuitively, Eq.~\ref{eq10} normalizes the reconstruction errors of different samples and then rescales the weights to ensure the summation of $ \bf M $ equals to $ P $.
A large prediction error means the corresponding sample performs as a hard one for the previously learned $ {\bf W}^{t-1}_\text{DSRR} $, which deserves more attention in the current learning.

In our experiments, a new discrete regressor is learned every 10 frames, and is updated to the previous model in a moving average manner: $ {\bf W}^{t}_\text{DSRR} = (1-\eta){\bf W}^{t-1}_\text{DSRR} + \eta{\bf W}_\text{DSRR} $.
An overview of the above model update process is presented in Algorithm~\ref{code1}.

\section{Experiments} \label{sec:experiments}

\subsection{Implementation Details}
In offline training, we freeze all the weights of the backbone network (ResNet-18 \cite{ResNet}) and adopt a multi-task training strategy to train the network $ \varphi(\cdot) $ and IoU predictor.
%
%
Note that the inputs of IoU predictor and ridge regression are different. Following ATOM \cite{ATOM}, the IoU predictor leverages the samples that have a certain overlap with the ground-truth box (at least 0.1). In contrast, our ridge regression branch utilizes the aforementioned positive and negative samples to learn the discriminative model.
The input RoI region is 5 times of the target size and is further resized to 288$ \times $288.
We utilize the training splits of LaSOT \cite{LaSOT}, TrackingNet \cite{2018trackingnet}, GOT-10k \cite{GOT10k}, and COCO \cite{COCO} for offline training.
The model is trained for 50 epochs with 1000 iterations per epoch and 36 image pairs per batch.
The ADAM optimizer \cite{ADAM} is employed with an initial learning rate of 0.01, and use a decay factor 0.2 for every 15 epochs.
The first-stage regressor uses ResNet-18 $ \mathrm{Block3} $ features as in ATOM, while the second-state regressor and the IoU predictor takes both $ \mathrm{Block3} $ and $ \mathrm{Block4} $ backbone features as input. 
In online tracking, to update the second-stage regressor, we collect 30 positive and 90 hard negative samples per reliable frame, and maintain a buffer for the last 30 frames.
The learning rate $ \eta $ of the second stage is 0.2. 
The reliability thresholds $ \tau_1 $ and $ \tau_2 $ are set to 0.25 and 0.4, respectively.

We denote our \underline{CA}scaded \underline{RE}gression method as {CARE} in the following experiments.
Our tracker is implemented in Python using PyTorch, and operates about 25 frames per second (FPS) on a single Nvidia GTX 1080Ti GPU.
We evaluate our method on each benchmark 3 times and report the average performance.

\subsection{Ablation Experiments}

We utilize the OTB-2015 \cite{OTB-2015}, UAV123 \cite{UAV123}, and LaSOT testing set \cite{LaSOT}, with total 503 videos, to comprehensively verify the effectiveness of our framework.

{\flushleft \bf Cascaded Framework.}
In Table~\ref{table:ablation study}, we compare the performance of each single stage and their cascaded combination. 
Note that we draw 512 samples per frame if the second stage is tested alone, aiming to obtain satisfactory performance. 
From Table~\ref{table:ablation study}, we can observe that the first and second stages almost perform identically. 
The main reason is that the convolutional regression is not discriminative enough, while the discrete sampling strategy fails to well cover a large search region. 
By combining them in a cascaded manner, superior performance can be obtained.
For example, on OTB-2015, our final cascaded tracker outperforms the first and second stages by 3.0\% and 3.1\%, respectively.
On the recent large-scale dataset LaSOT, our final framework surpasses the first stage by 3.0\% in AUC. 
Note that the first stage in our framework is adopted from the ATOM, which already achieves a high performance level on various challenging datasets.
Under the same backbone network and bounding box regression manner (i.e., IoUNet), our performance gains can be attributed to the superior discrimination capability of our cascaded framework.
As for the tracking speed, with the abundant candidates (512 samples per frame), the second-stage regressor is less efficient than the first stage. 
In contrast, our cascaded framework achieves a balanced speed and outstanding performance, which only slightly reduces the first-stage efficiency but notably outperforms it in tracking accuracy.

\setlength{\tabcolsep}{2pt}
\begin{table}[t]
	\scriptsize
	\begin{center}
		\caption{Analysis of each component in our method. We first compare the performance of the first stage, second stage and their cascaded combination. Then, we analyze the second stage by adding re-detection mechanism and adaptive ridge regression (ADRR). The performance is verified on OTB-2015, UAV123, and LaSOT in terms of the area-under-curve (AUC) score of success plot.} \label{table:ablation study}	
		\begin{tabular*}{8.5 cm} {@{\extracolsep{\fill}}lcccc|ccc|c}
			\hline
			&First &Second & Re-detection & ADRR & OTB-2015 & UAV123 &LaSOT & Speed \\
			&Stage &Stage &  & & \cite{OTB-2015} & \cite{UAV123} &\cite{LaSOT} & FPS \\
			\hline
			&$\checkmark$ & & & &67.5 &63.3 &51.7 &{\bf 30}\\
			& &$\checkmark$  & & &68.0 &62.1 &49.3 &20\\
			&$\checkmark$ &$\checkmark$  & &&69.2 &64.4 &53.7 &27\\
			\hline
			&$\checkmark$ &$\checkmark$  &$\checkmark$ &&69.5 &65.0 &54.1 &25\\
			&$\checkmark$ &$\checkmark$  &$\checkmark$ &$\checkmark$&{\bf 70.5} &{\bf 65.4} &{\bf 54.7} &25\\
			\hline
		\end{tabular*}
	\end{center}
\end{table}

{\flushleft \bf Target Re-detection.}
As discussed in Section~\ref{online tracking}, our second-stage regressor also acts as a re-detection module due to its high discrimination. 
As shown in Table~\ref{table:ablation study}, with additional performance improvements, the re-detection mechanism further exploits the potential of the second stage.

{\flushleft \bf Adaptive Ridge Regression.}
Online model update plays a vital role in our framework. Based on the collected hard samples in the memory buffer, to better concentrate on the valuable ones, we propose an adaptive ridge regression that dynamically reweighs the training samples.
As illustrated in Table~\ref{table:ablation study}, our adaptive ridge regression (ADRR) steadily improves the tracking accuracy.
Besides, it is worth mentioning that our ADRR is extremely efficient with a negligible computational cost.

\subsection{Comparison with State-of-the-art Methods}
We compare our proposed CARE tracker with the recent state-of-the-art trackers on 11 challenging tracking benchmarks including OTB-2013 \cite{OTB-2013}, OTB-2015 \cite{OTB-2015}, UAV123 \cite{UAV123}, LaSOT \cite{LaSOT}, VOT2018 \cite{VOT2018}, VOT2019 \cite{VOT2019}, TrackingNet \cite{2018trackingnet}, Temple-Color \cite{TempleColor128}, UAV20L \cite{UAV123}, Need for Speed \cite{NFSdataset}, and OxUvA \cite{2018longtermBenchmark}.

\begin{figure}[t]
	\centering
	\includegraphics[width=4.3cm]{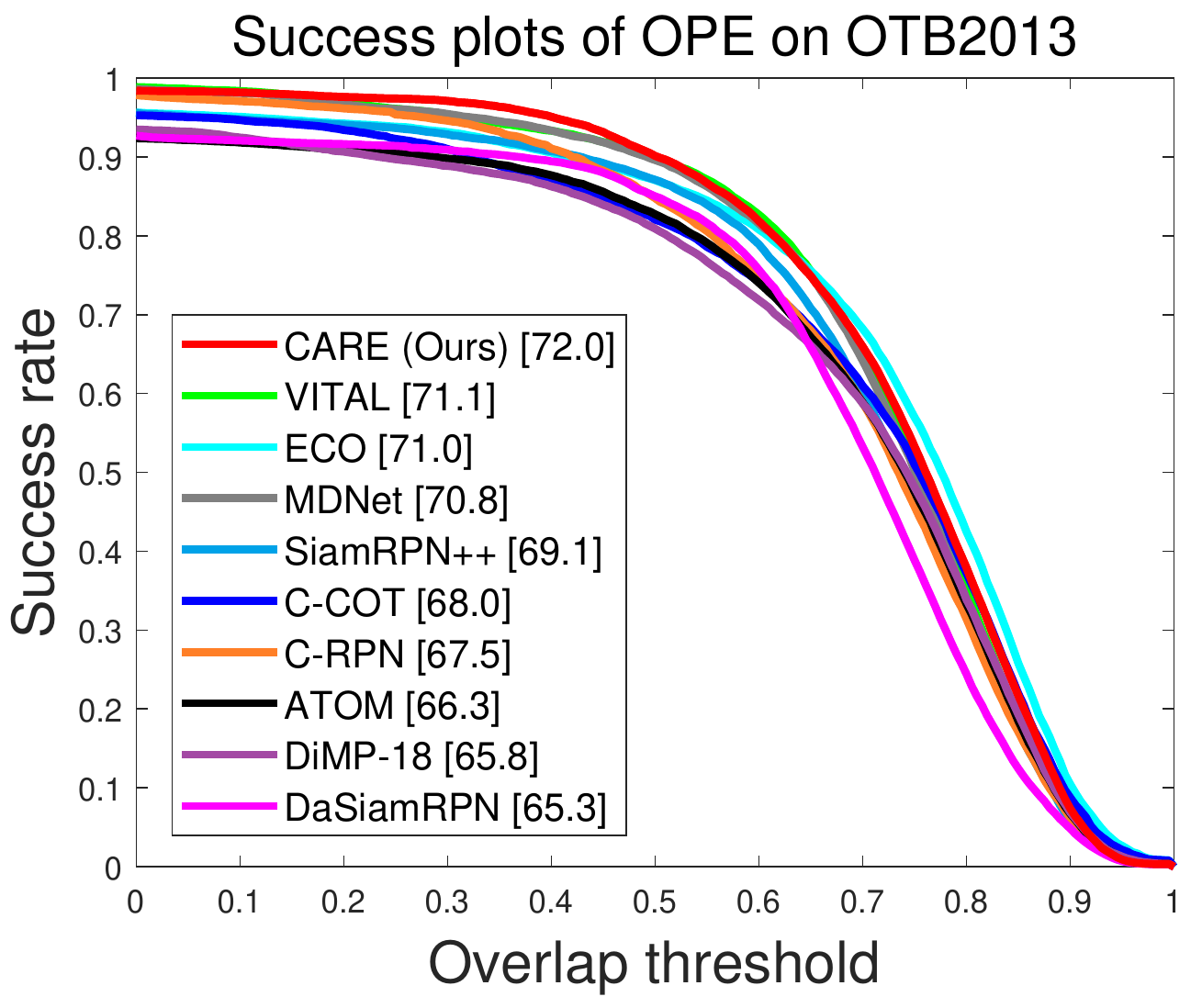}
	\includegraphics[width=4.3cm]{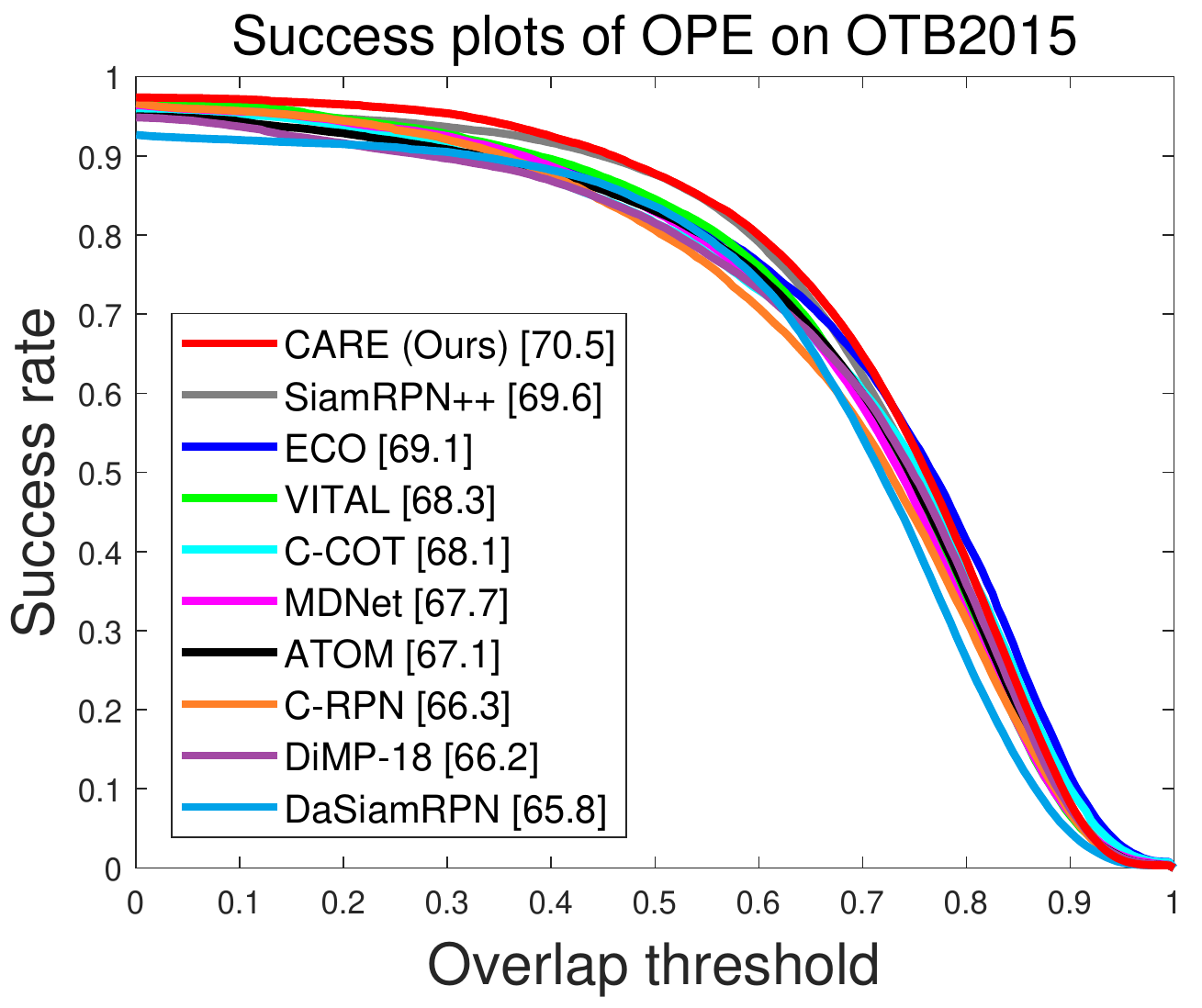}
	\caption{Success plots on the OTB-2013 \cite{OTB-2013} (left) and OTB-2015 \cite{OTB-2015} (right) datasets. The legend shows the AUC score. The proposed CARE method outperforms all the comparison trackers.} 
	\label{fig:otb-50-100} 
\end{figure}

{\flushleft \bf OTB-2013 \cite{OTB-2013}.} OTB-2013 is a widely evaluated tracking dataset with 50 videos. Figure \ref{fig:otb-50-100} (left) shows the success plot on the OTB-2013.
On this dataset, our method achieves an AUC score of 72.0\%, outperforming all previous state-of-the-art trackers such as VITAL \cite{VITAL} and ECO \cite{ECO}.
Note that the top-performing trackers on this benchmark cannot operate at a real-time speed, e.g., the speeds of VITAL and MDNet are only 1 FPS by using fully-connected layers for candidate classification, while ours is real-time since our closed-form regressor is free of gradient back-propagation.
Compared with other state-of-the-art trackers with the same ResNet backbone (e.g., ATOM), our approach exhibits competitive efficiency with a speed of about 25 FPS.

{\flushleft \bf OTB-2015 \cite{OTB-2015}.} 
OTB-2015 benchmark extends OTB-2013 with additional 50 videos, resulting in 100 videos in total.
Figure \ref{fig:otb-50-100} (right) shows the success plot over 100 videos on the OTB-2015.
Our method achieves an AUC score of 70.5\% on this benchmark, surpassing the recently proposed SiamRPN++ \cite{siamrpn++}, ECO \cite{ECO}, and VITAL \cite{VITAL} trackers. 
Compared with the recent single-stage regression trackers such as ATOM \cite{ATOM} and DiMP-18 \cite{DiMP}, our CARE method outperforms them by 3.4\% and 4.3\% in terms of AUC score, respectively.
Note that DiMP-18 is the recently proposed regression method with discriminative model learning, which represents the state-of-the-art performance on several datasets.

\begin{figure}[t]
	\centering
	\includegraphics[width=4.3cm]{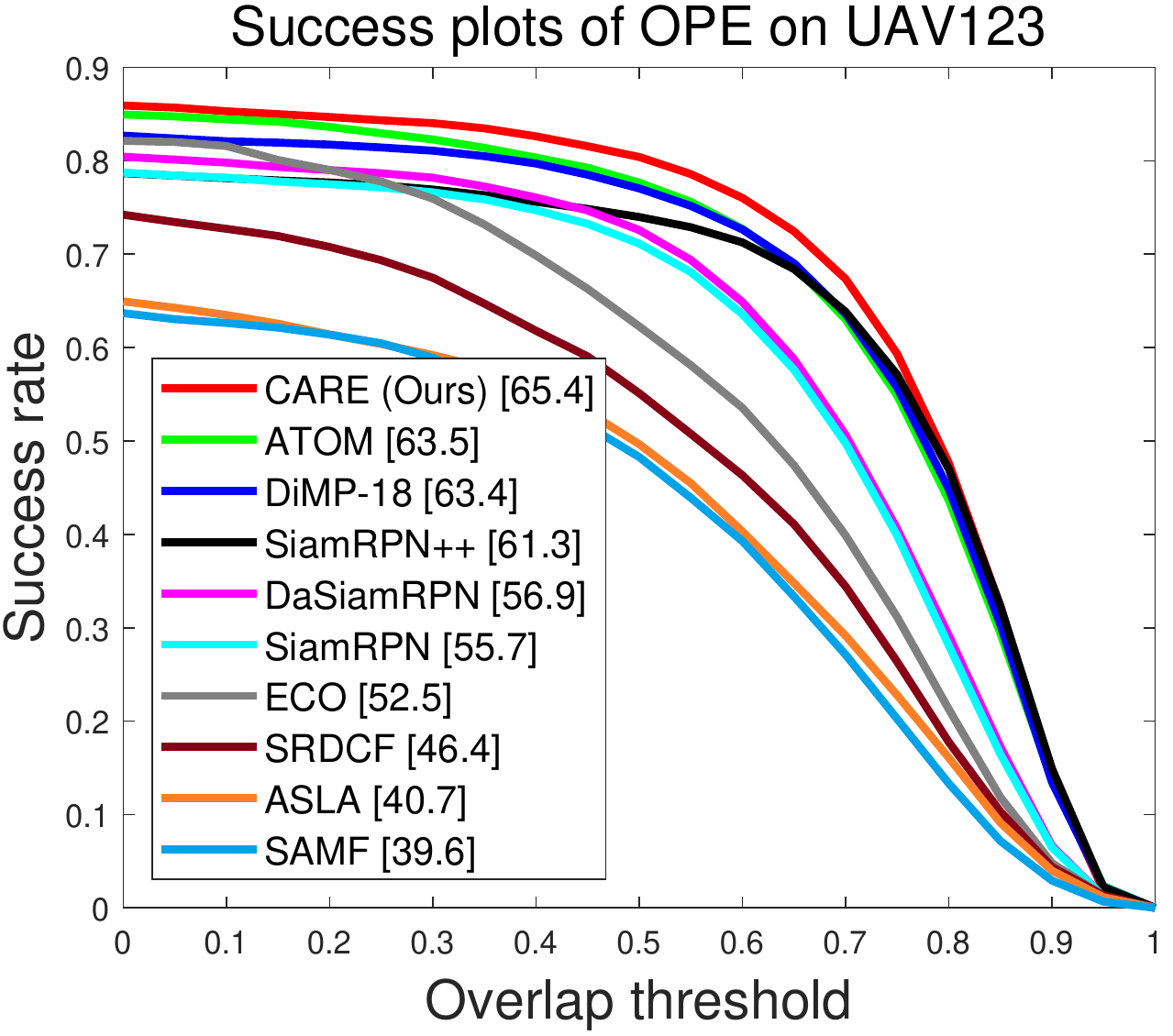}
	\includegraphics[width=4.3cm]{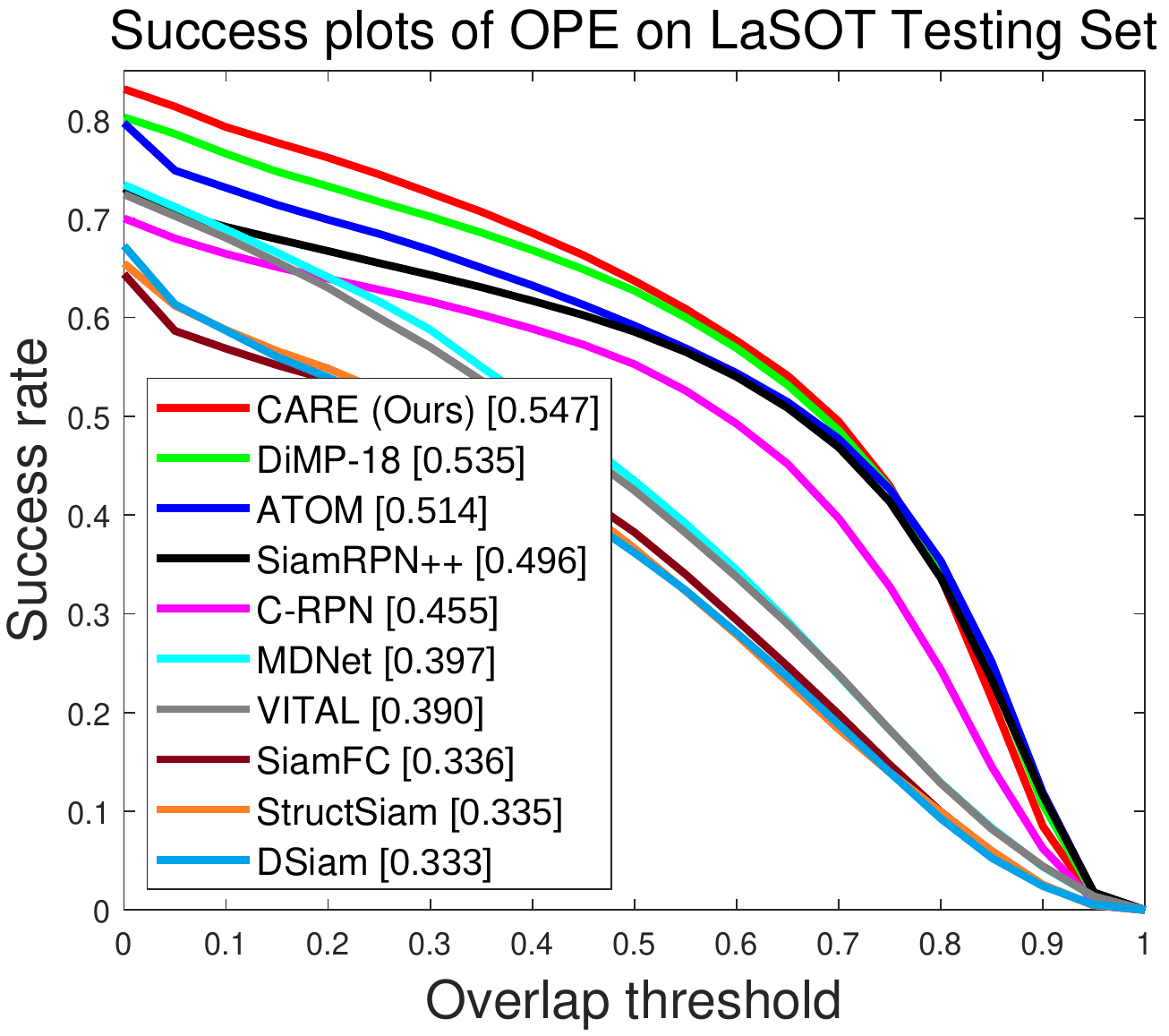}
	\caption{Success plots on the UAV123 \cite{UAV123} (left) and LaSOT \cite{LaSOT} (right) datasets. The legend shows the AUC score. The proposed CARE method outperforms all the comparison trackers.}
	\label{fig:uav-lasot} 
\end{figure}

{\flushleft \bf UAV123 \cite{UAV123}.} 
This dataset includes 123 aerial videos collected by a low-attitude UAV platform. 
Therefore, UAV123 focuses on evaluating visual trackers in the UAV scenarios with small and fast-moving targets. 
Figure \ref{fig:uav-lasot} (left) illustrates the success plot of the state-of-the-art trackers including SiamRPN++, ATOM, and DiMP-18. 
Compared with the recent remarkable approaches, our method achieves the best result. 
Especially, our approach shows an AUC score of 65.4\%, outperforming SiamRPN++, ATOM, and DiMP-18 by 4.1\%, 1.9\%, and 2.0\% AUC score, respectively. 
Since our approach is equipped with the same backbone network and IoU predictor compared with ATOM and DiMP-18, our performance advantage verifies the superiority of the proposed cascaded tracking framework.

{\flushleft \bf LaSOT \cite{LaSOT}.} LaSOT is a recent large-scale tracking dataset including 1200 videos, which is more challenging than the previous short-term benchmarks with an average of 2500 frames per video. 
We evaluate our approach on the test set of 280 videos. 
Except for the top-performing trackers like MDNet and VITAL on this dataset, we also include the recent C-RPN \cite{CRPN}, SiamRPN++, ATOM, and DiMP-18 for comparison.
The success plot on LaSOT is shown in Figure~\ref{fig:uav-lasot} (right).
On this dataset, our approach achieves an AUC score of 54.7\%, outperforming the previous best method on this benchmark (i.e., MDNet) by a considerable margin of 15.0\% AUC score.
Compared with the recent C-RPN, SiamRPN++, ATOM, and DiMP-18, our CARE surpasses them by 9.2\%, 5.1\%, 3.3\%, and 1.2\% in AUC, respectively.

\begin{figure}[t]
	\centering
	\includegraphics[width=8.6cm]{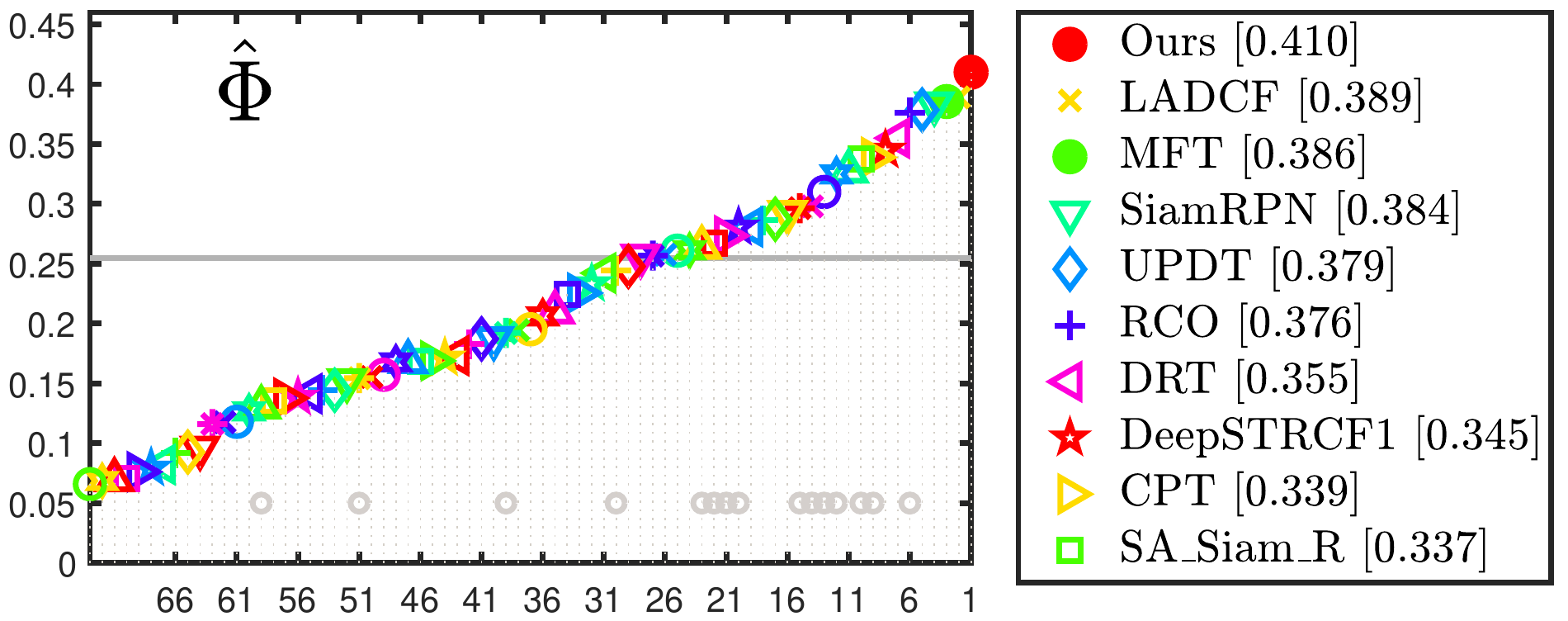}
	\caption{Expected average overlap (EAO) graph with trackers ranked from right to left. Our method obviously outperforms all the participants on the VOT2018 \cite{VOT2018}.}
	\label{fig:vot} 
\end{figure}

\setlength{\tabcolsep}{2pt}
\begin{table}[t]
	\scriptsize
	\begin{center}
		\caption{Comparison with recent state-of-the-art trackers on the VOT2018 \cite{VOT2018} in terms of robustness (R), accuracy (A), and expected average overlap (EAO).} \label{table:vot}	 
		\begin{tabular*}{8.6 cm} {@{\extracolsep{\fill}}lcccccccc}
			\hline
			~ & SPM &C-RPN &DWSiam &SiamMask & SiamRPN++ &ATOM &DiMP-18 &{\bf CARE} \\
			~ & \cite{SPM} &\cite{CRPN} &\cite{deeperwiderSiamFC} &\cite{SiamMask} &\cite{siamrpn++} &\cite{ATOM} &\cite{DiMP} & \\
			\hline
			~R  &0.30 &- &- &0.276 &0.234 &0.204 &{\bf \color{red} 0.182} &{\bf \color{blue} 0.201}\\
			~A  &0.58 &- &- &{\bf \color{red} 0.609} &{\bf \color{blue} 0.600} &0.590 &0.594 &0.597\\
			~EAO   &0.338 &0.289 &0.301 &0.380 &{\bf \color{red}0.414} &0.401 &0.402 &{\bf \color{blue} 0.410} \\
			\hline
			~FPS &110 &32 &150 &55 &35 &30 &46 &25 \\
			\hline
		\end{tabular*}
	\end{center}
\end{table}

\setlength{\tabcolsep}{2pt}
\begin{table}[t]
	\scriptsize
	\begin{center}
		\caption{Comparison with recent state-of-the-art trackers on the VOT2019 \cite{VOT2019} in terms of robustness (R), accuracy (A), and expected average overlap (EAO).} \label{table:vot19}	 
		\begin{tabular*}{8.5 cm} {@{\extracolsep{\fill}}lccccccc}
			\hline
			~ & SPM  &SiamMask &SiamMask-E & SiamRPN++ &ATOM &SiamDW &{\bf CARE} \\
			~ & \cite{SPM}  &\cite{SiamMask} &\cite{chen2019SiamMaskRotated} &\cite{siamrpn++} &\cite{ATOM} &\cite{deeperwiderSiamFC} & \\
			\hline
			~R   &0.507 &0.461 &0.487 &0.482 &{\bf \color{blue} 0.411} &0.467 &{\bf \color{red} 0.343} \\
			~A   &0.577 &0.594 &{\bf \color{red} 0.652} &0.599 &{\bf \color{blue} 0.603} &0.600 &0.601 \\
			~EAO  &0.275 &0.287 &0.309 &0.285 &0.292 &{\bf \color{blue} 0.299} &{\bf \color{red} 0.323} \\
			\hline
			~FPS  &110 &55 &50 &35 &30 &- &25 \\
			\hline
		\end{tabular*}
	\end{center}
\end{table}

\begin{figure*}
	\centering
	\includegraphics[width=4.4cm]{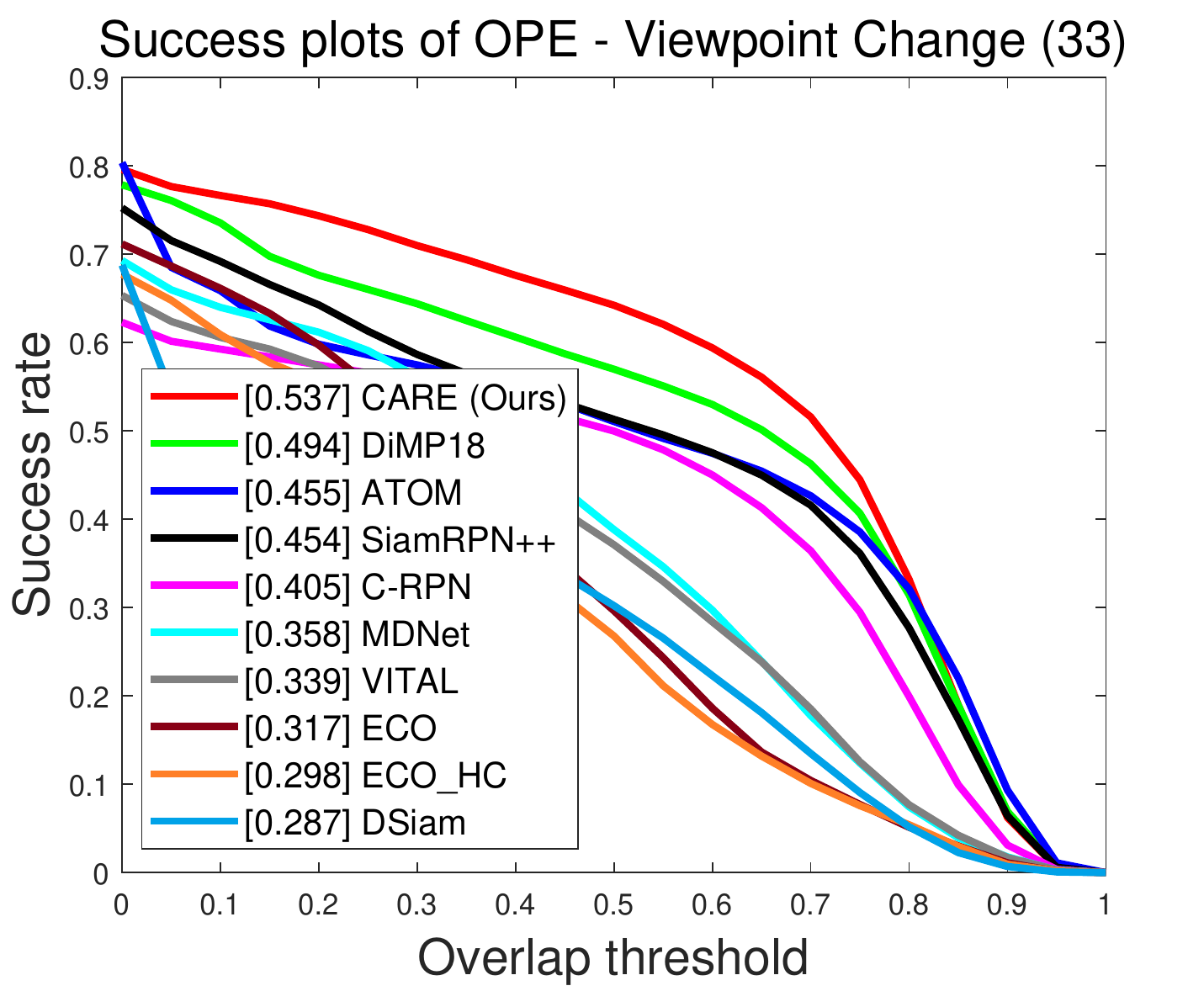}
	\includegraphics[width=4.4cm]{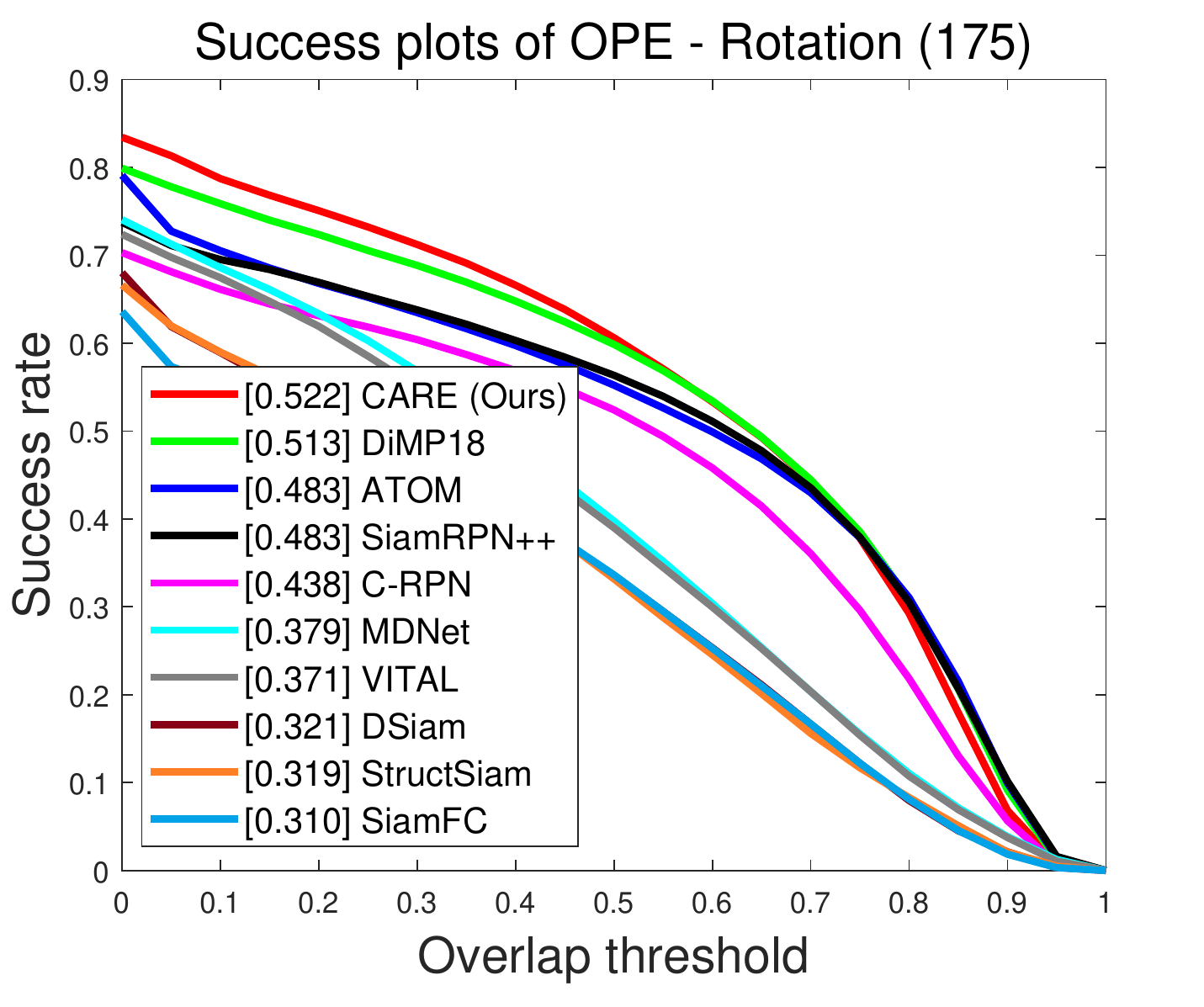}
	\includegraphics[width=4.4cm]{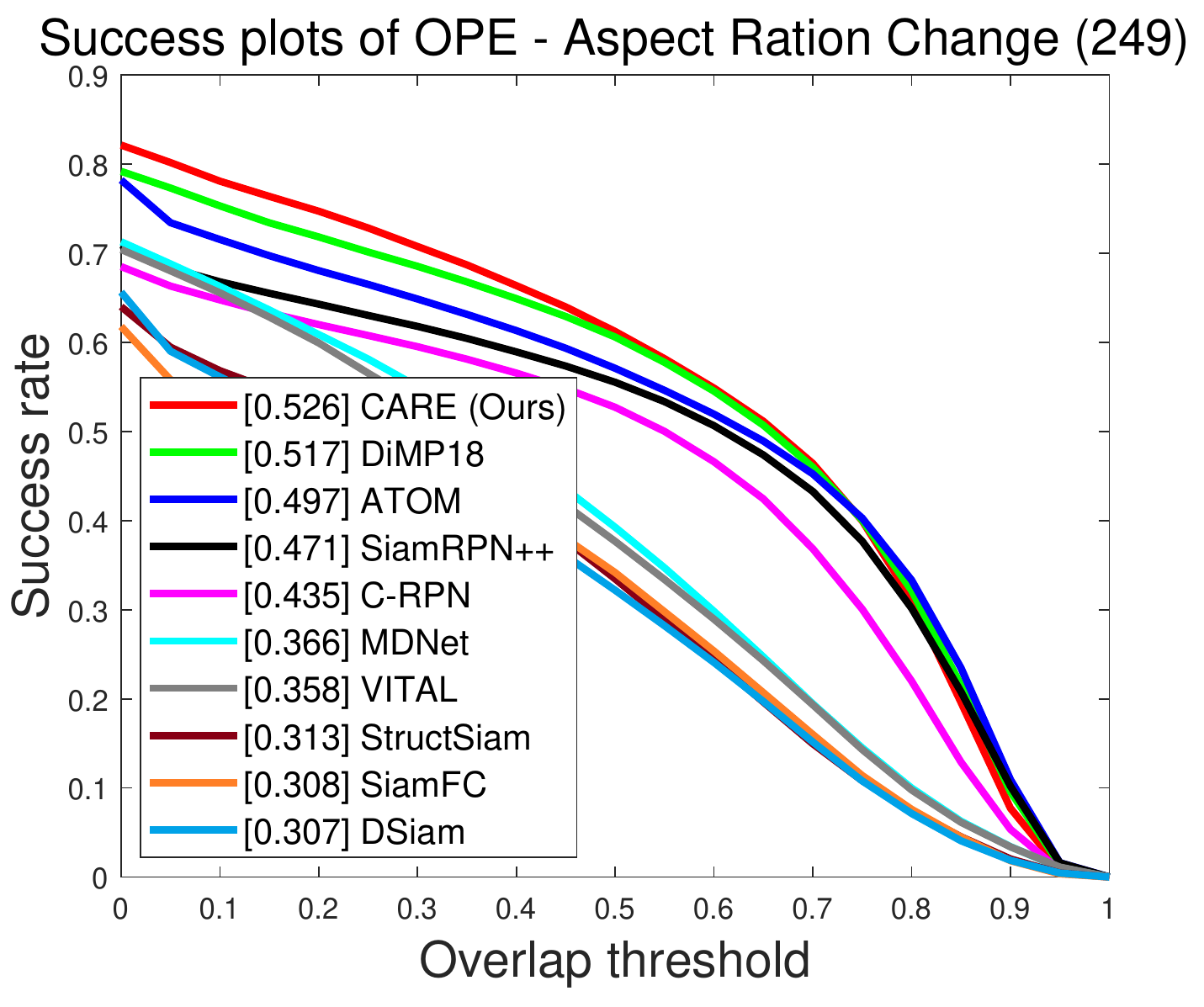}
	\includegraphics[width=4.4cm]{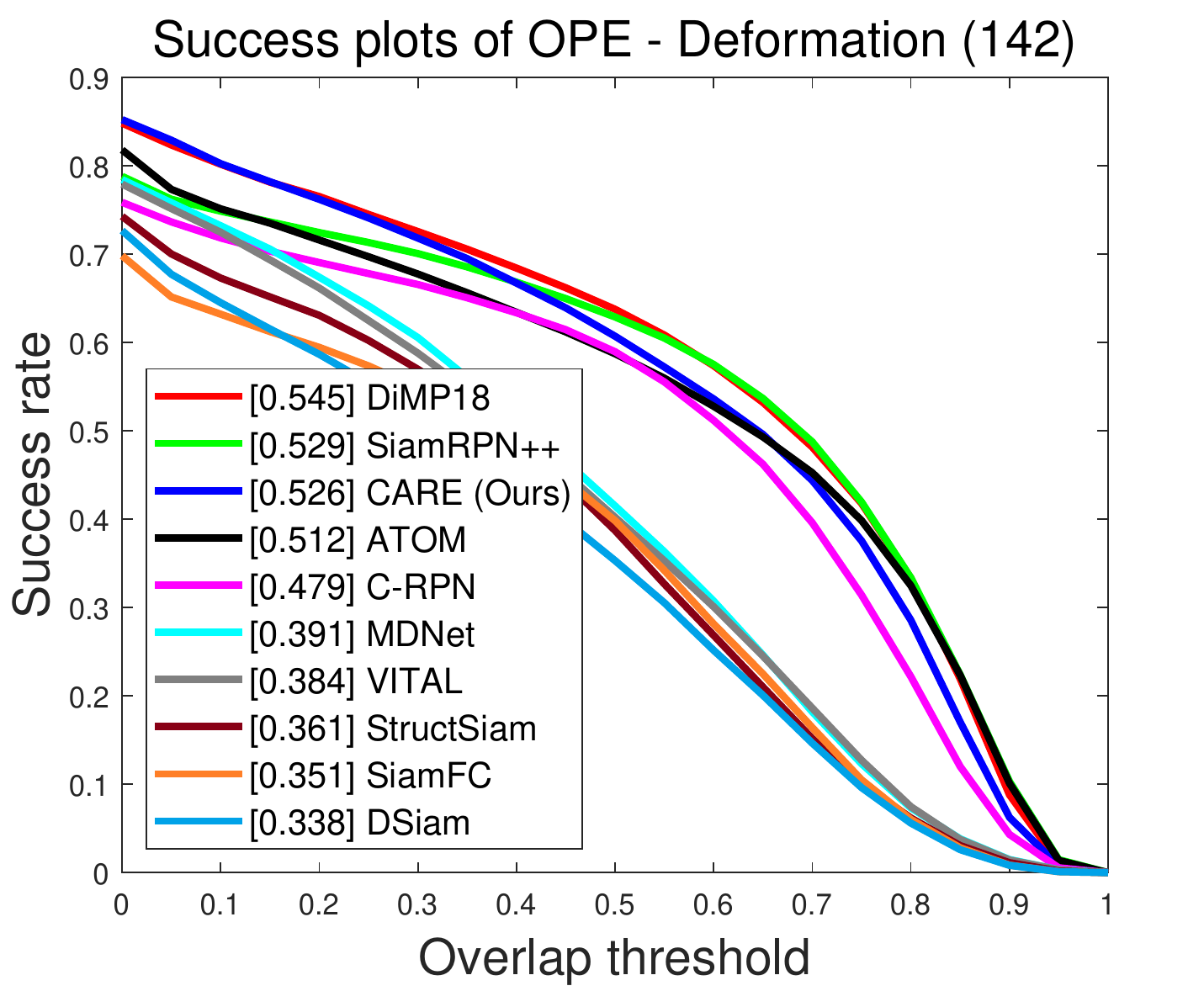}
	\includegraphics[width=4.4cm]{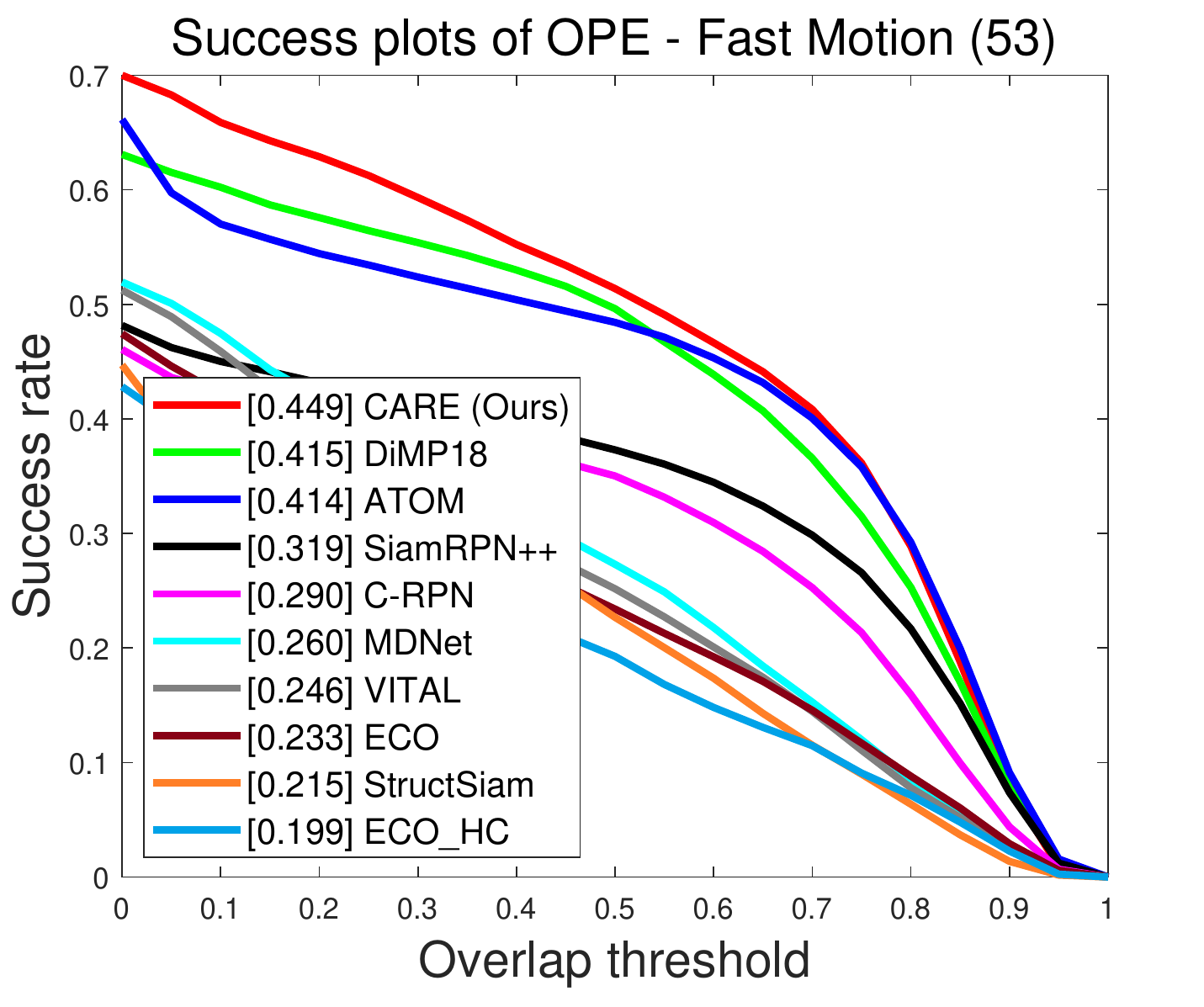}
	\includegraphics[width=4.4cm]{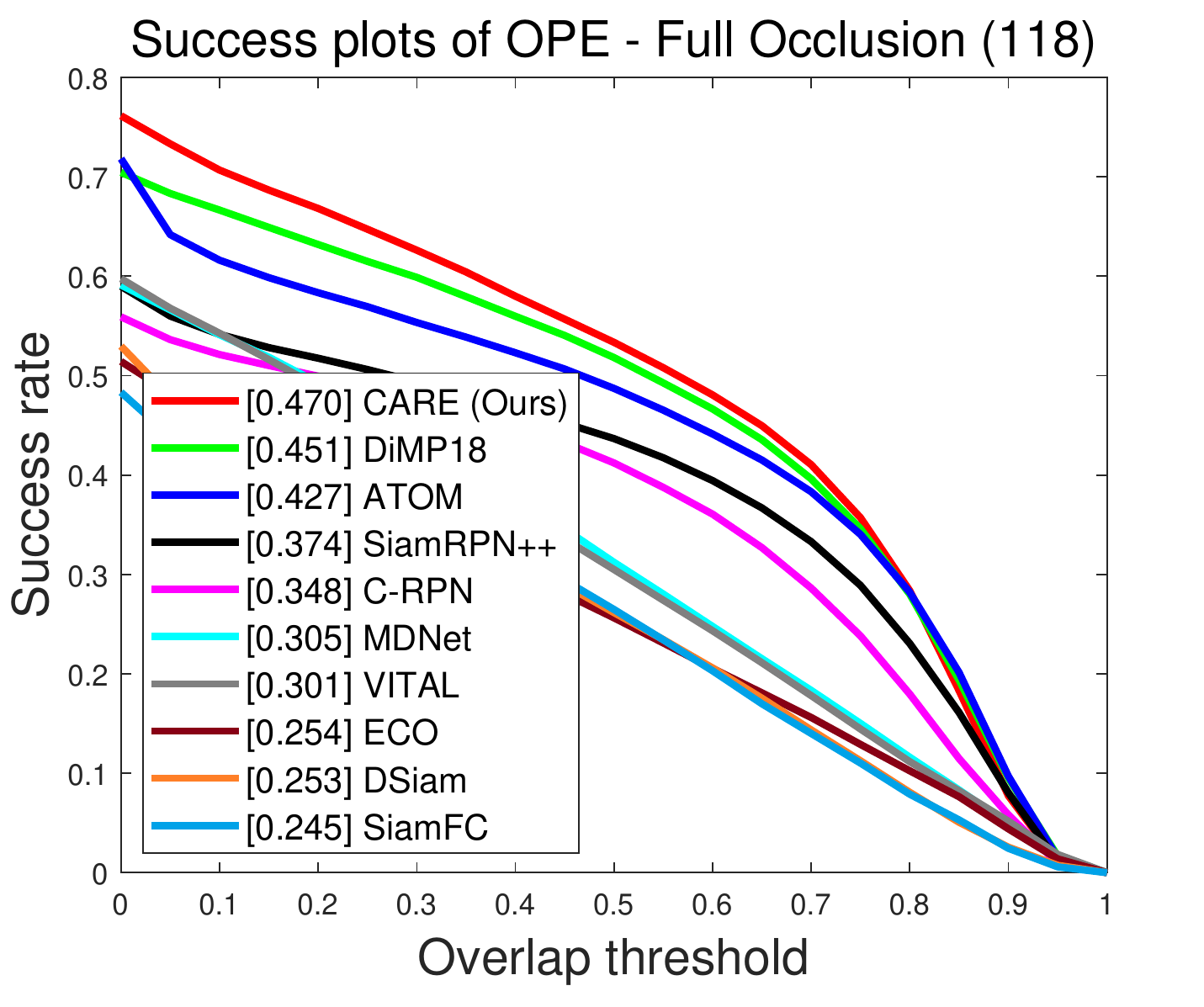}
	\includegraphics[width=4.4cm]{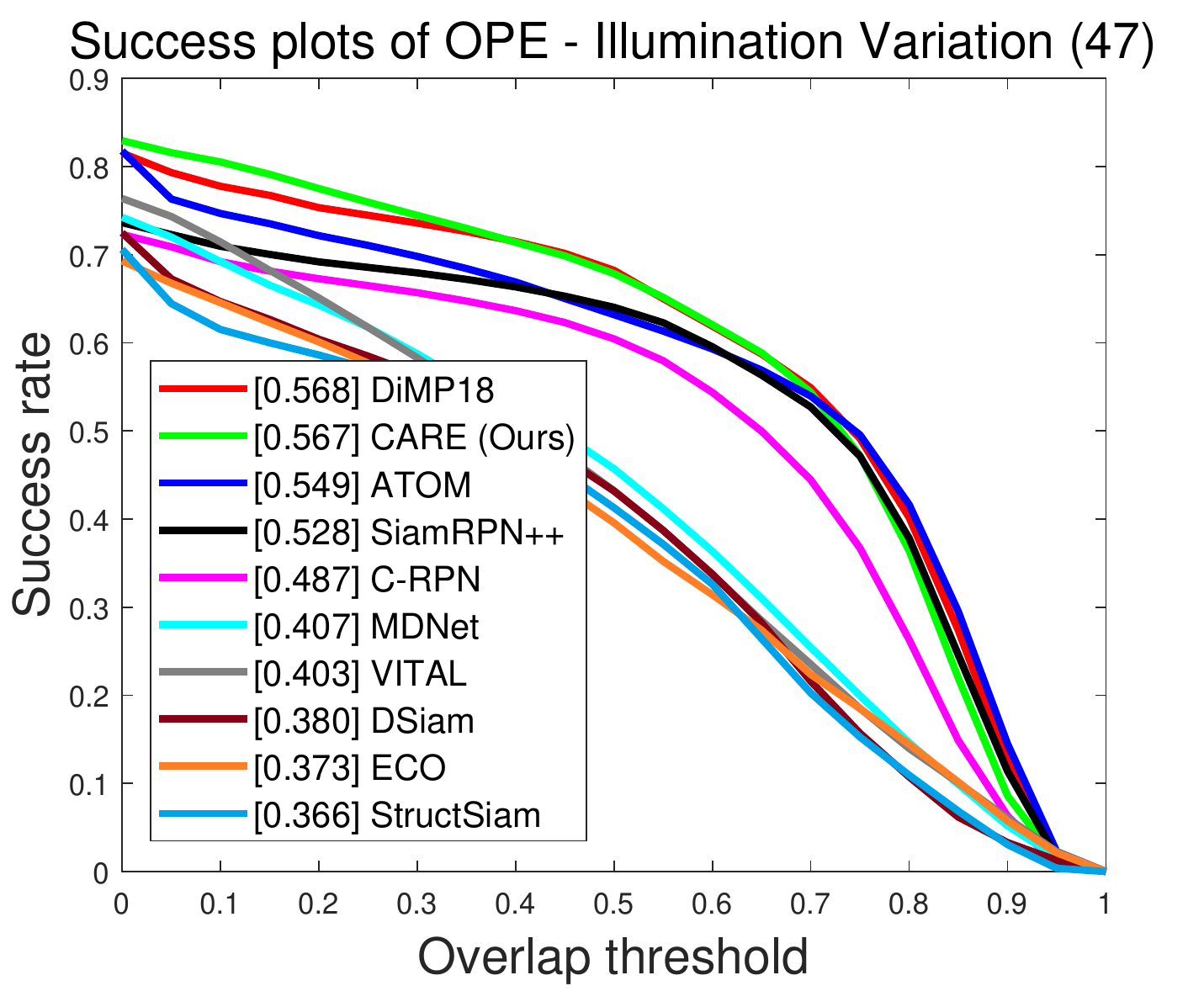}
	\includegraphics[width=4.4cm]{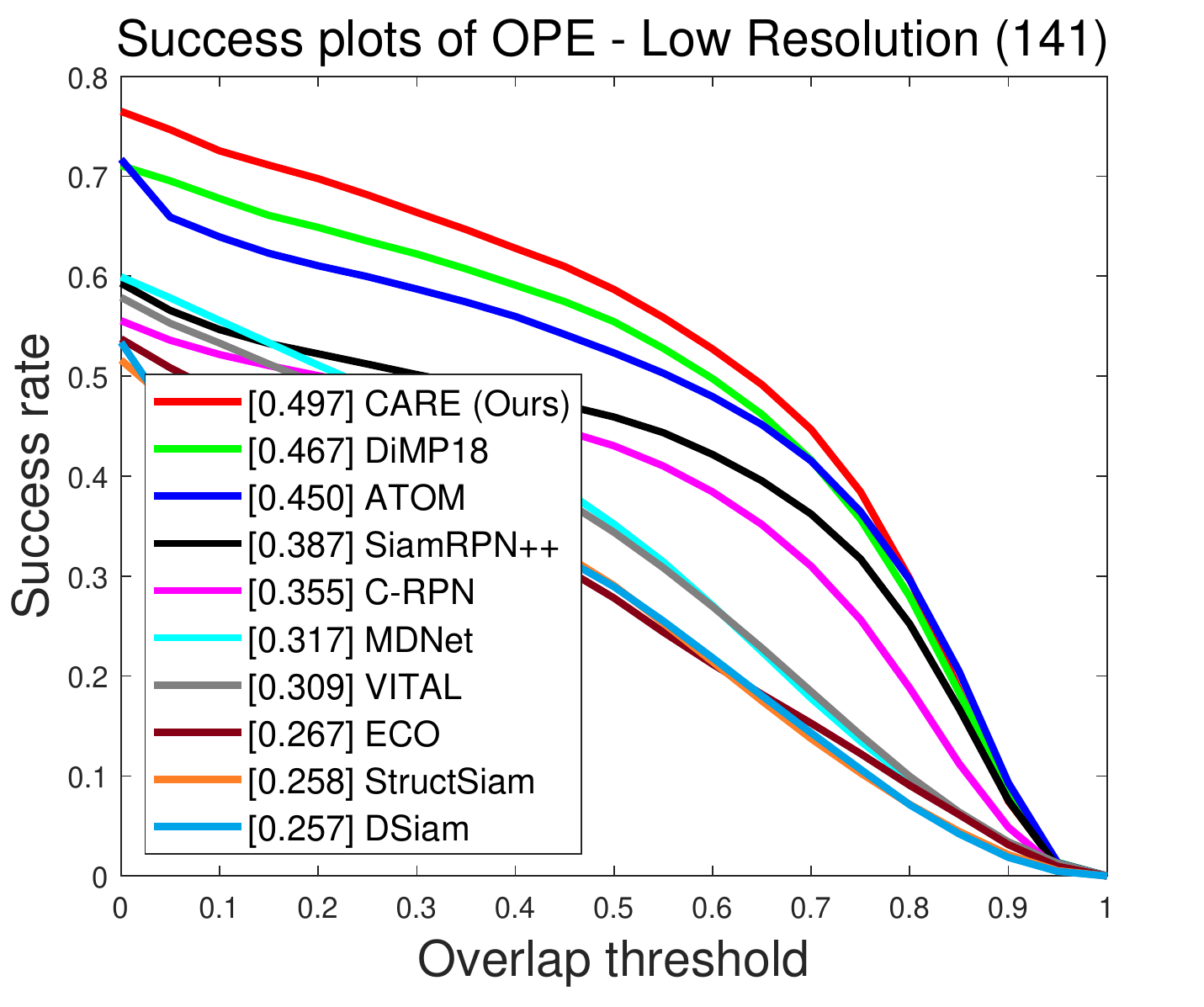}
	\includegraphics[width=4.4cm]{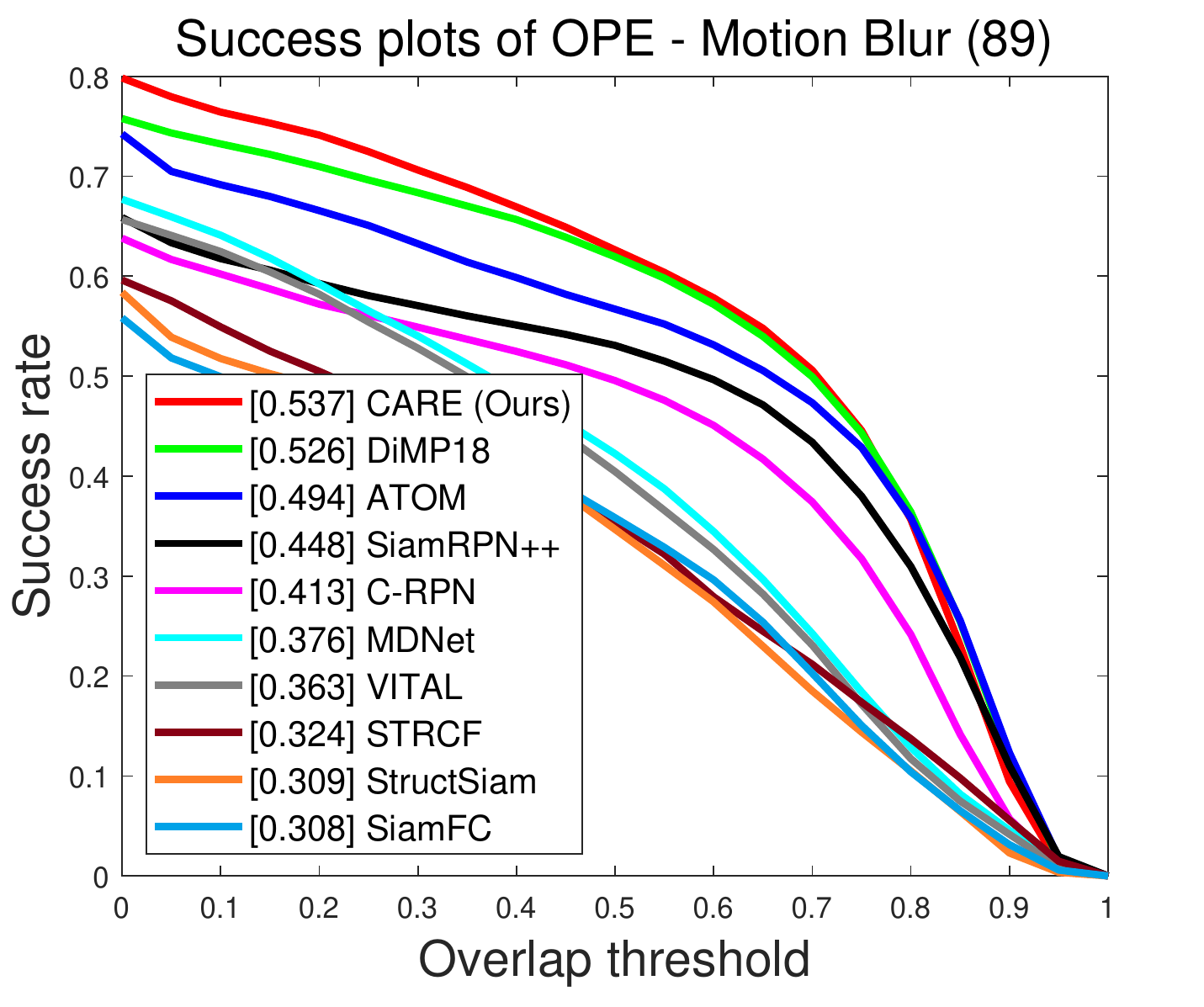}
	\includegraphics[width=4.4cm]{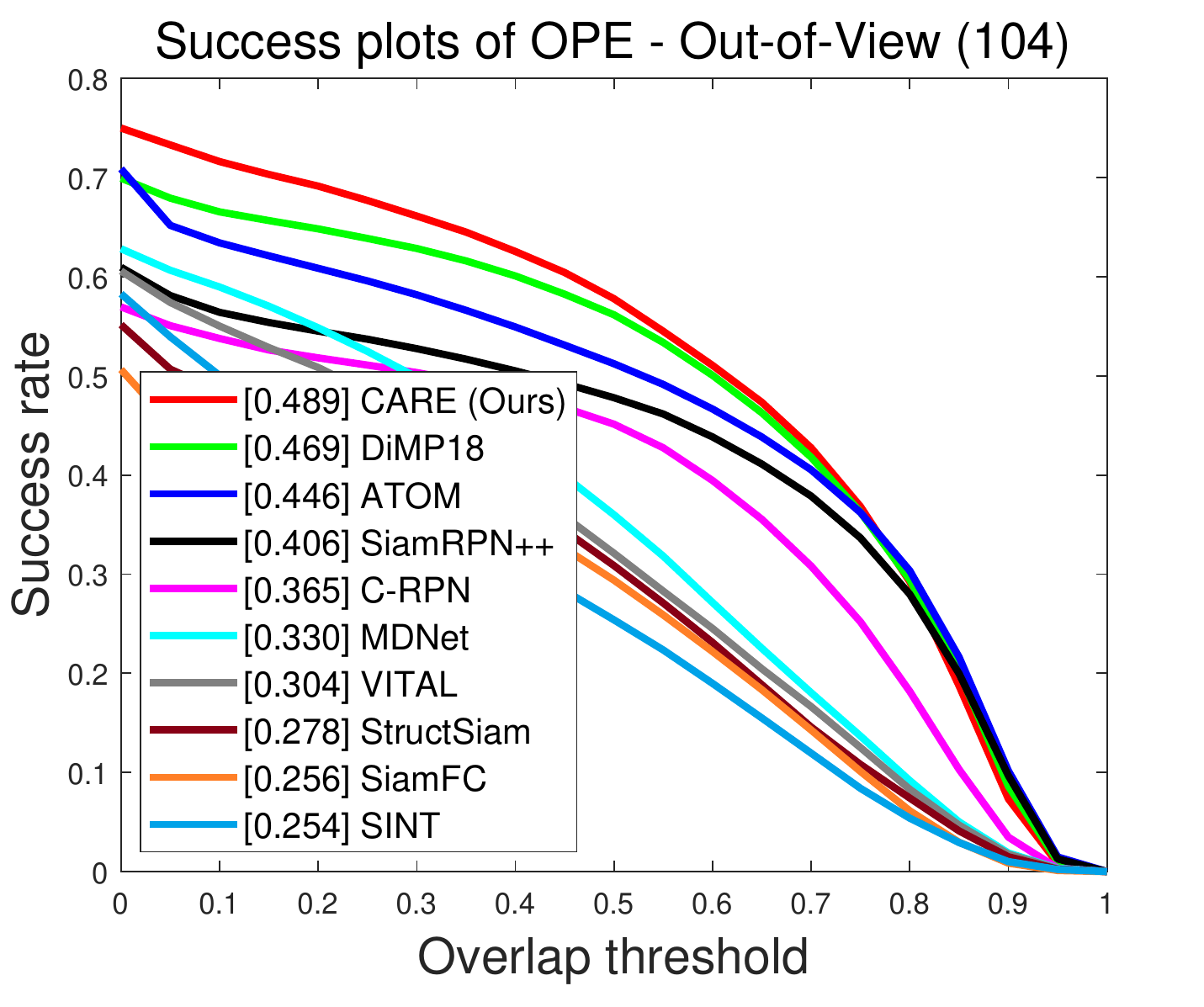}
	\includegraphics[width=4.4cm]{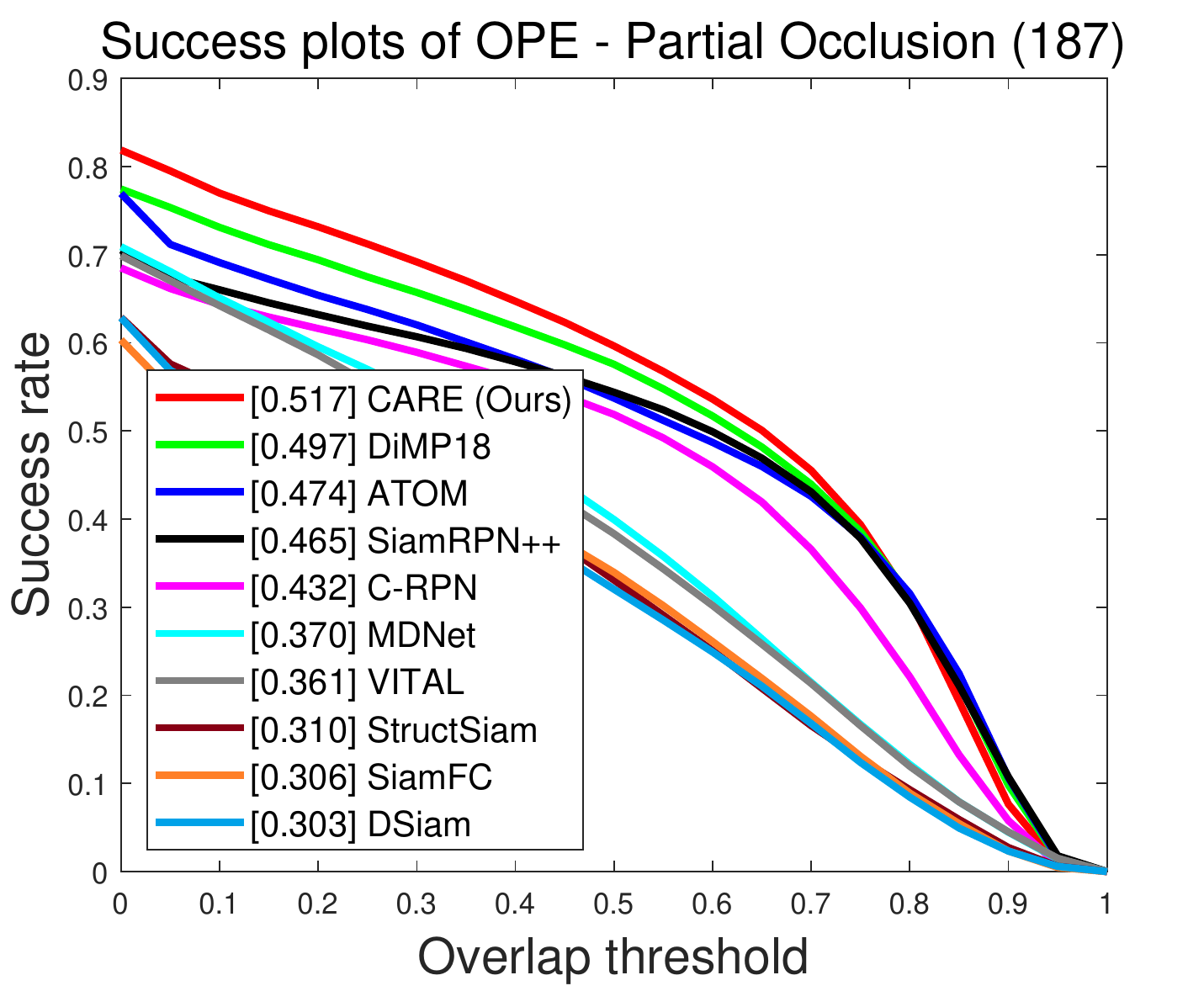}
	\includegraphics[width=4.4cm]{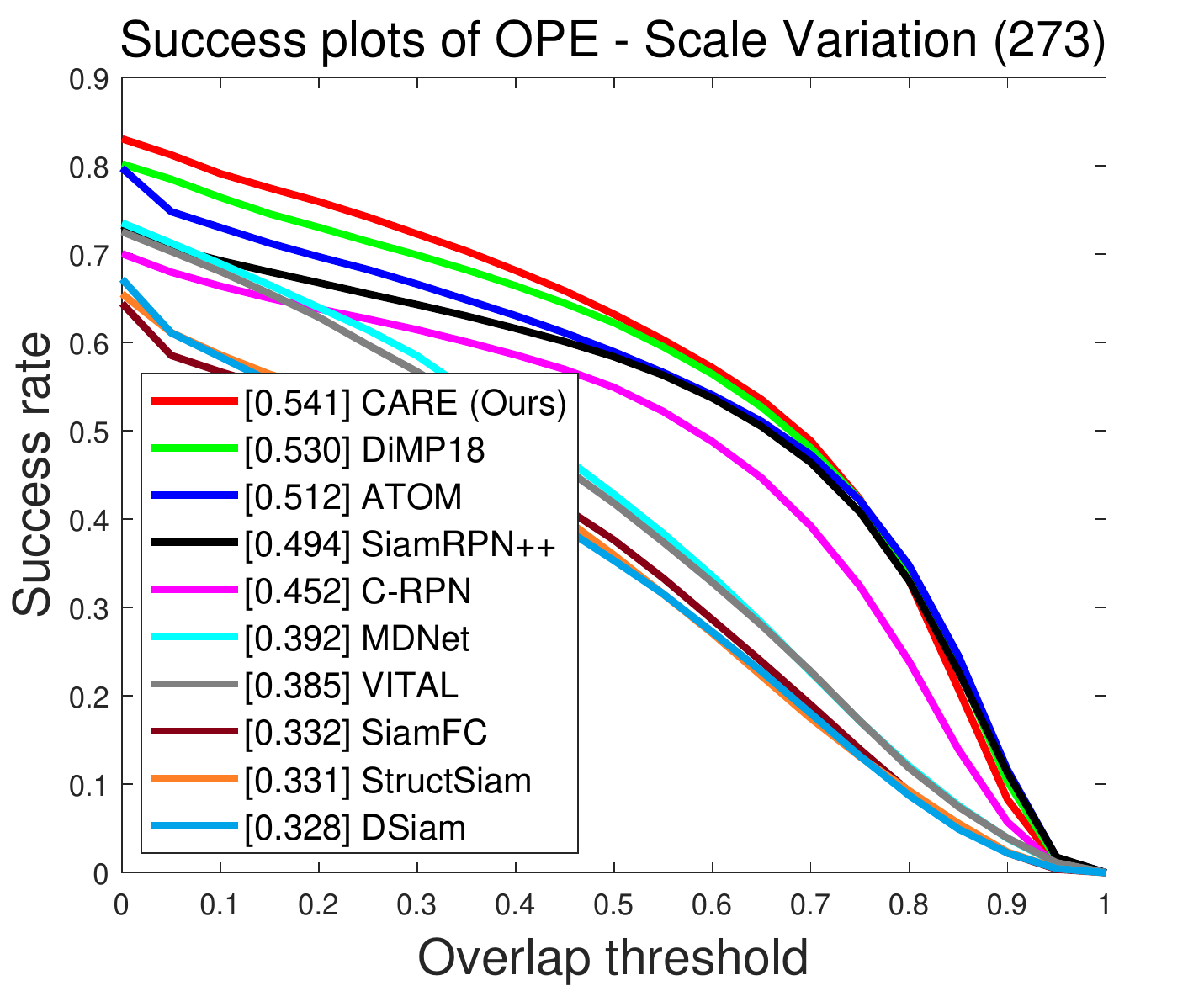}
	\caption{Attribute-based evaluation on the LaSOT benchmark \cite{LaSOT} including viewpoint change, rotation, aspect ration change, deformation, fast motion, full occlusion, illumination variation, low resolution, motion blur, out-of-view, partial occlusion, and scale variation. The legend shows the AUC score of the comaprison tracker.}
	\label{fig:lasot-attribute} 
\end{figure*}

In Figure~\ref{fig:lasot-attribute}, we further provide the attribute evaluation on the LaSOT benchmark \cite{LaSOT}.
On this large-scale dataset, our approach shows good results on fast motion, out-of-view, and viewpoint change.
On the above attributes, our method even surpasses the recently remarkable DiMP-18 tracker \cite{DiMP} by a large margin, which can be attributed to the strong discrimination of our second stage.
Our second-stage regressor further checks the ambiguous candidates and serves as a re-detection module, which significantly improves the tracking performance on the challenging scenarios such as fast motion, out-of-view, and target occlusion. 
Besides, our method outperforms its baseline method ATOM \cite{ATOM} in all attributes on the LaSOT dataset.
In particular, our method significantly outperforms ATOM in viewpoint change, low resolution, and partial occlusion by 8.2\%, 4.7\%, and 4.3\%, respectively, which demonstrates the effectiveness of our second stage for cascaded verification. 
Since our framework mainly focuses on the target re-identification and re-detection, in the attributes such as aspect ratio change and scale variation, our method is less effective and slightly improves the baseline.

\setlength{\tabcolsep}{2pt}
\begin{table*}[t]
	\scriptsize
	\begin{center}
		\caption{Comparison with state-of-the-art trackers on the TrackingNet \cite{2018trackingnet} test set in terms of precision, normalized precision, and success (AUC score of the success plot).} \label{table:trackingnet}	
		\begin{tabular*}{18.0 cm} {@{\extracolsep{\fill}}lcccccccccccccccc}
			\hline
			~ &BACF &Staple &Staple-CA &CSR-DCF &ECOhc &ECO &SiamFC &CFNet &MDNet &UPDT &DaSiamRPN &SPM &C-RPN & ATOM &DiMP-18 &{\bf CARE}\\
			~ &\cite{BACF} &\cite{Staple} &\cite{Context-AwareCorrelationFilter} &\cite{CSR-DCF} &\cite{ECO} &\cite{ECO} &\cite{SiamFC} &\cite{CFNet} &\cite{MDNet} &\cite{UPDT} &\cite{DaSiamRPN} &\cite{SPM} &\cite{CRPN}  &\cite{ATOM} &\cite{DiMP} &\\
			\hline
			~Precision &46.1 &47.0 &46.8 &48.0 & 47.6 &49.2 &53.3  &53.3 &56.5 &55.7 &59.1 &66.1 &61.9 &64.8 &{\bf \color{blue} 66.6} &{\bf \color{red} 66.7} \\
			~Norm. Prec. &58.0 &60.3 &60.5 &62.2 &60.8  &61.8 &66.3  &65.4  &73.3 &70.2 &73.3 &77.8 &74.6  &77.1 &{\bf \color{blue} 78.5} &{\bf \color{red} 79.0} \\
			~Success &52.3 &52.8 & 52.9 &53.4 &54.1 &55.4 &57.1  &57.8  &63.8 &61.1 &63.8 &71.2 &66.9 &70.3 &{\bf \color{red} 72.3} &{\bf \color{blue} 71.8}\\
			\hline
			~Speed (FPS) &35 &70 &55 &18 &45 &8 &86 &55 &1 &- &{\bf \color{red} 160} &{\bf \color{blue} 110} &32 &30 &46 &25\\
			\hline
		\end{tabular*}
	\end{center}
\end{table*}

\setlength{\tabcolsep}{2pt}
\begin{table*}[t]
	\scriptsize
	\begin{center}
		\caption{Comparison with state-of-the-art trackers on the UAV-20L \cite{UAV123}, Temple Color \cite{TempleColor128}, and Need for Speed \cite{NFSdataset} datasets. The evaluation metric is the AUC score of the success plot.} \label{table:TC_NFS}	
		\begin{tabular*}{18.0 cm} {@{\extracolsep{\fill}}lccccccccccccccc}
			\hline
			~ &KCF &DSST &SRDCF &HCF &SiamFC &CFNet &ECOhc &MDNet &C-COT &ECO &SiamRPN &SiamRPN++ & ATOM &DiMP-18 &{\bf CARE}\\
			~ &\cite{KCF} &\cite{DSST}  &\cite{SRDCF} &\cite{HCF} &\cite{SiamFC} &\cite{CFNet} &\cite{ECO} &\cite{MDNet} &\cite{C-COT} &\cite{ECO} &\cite{SiamRPN} &\cite{siamrpn++} &\cite{ATOM} &\cite{DiMP} &\\
			\hline
			~UAV20L \cite{UAV123} &19.8 &27.0 &34.3 &- &39.9 &34.9 &- &- &- &43.5 &45.4 &56.1 &55.4 &{\bf \color{blue} 57.1} &{\bf \color{red} 60.3} \\
			~TC128 \cite{TempleColor128} &38.4 &40.6 &50.9 &48.2 &50.5 &45.6 &56.1 &56.3 &58.3 &59.7 &- &56.2 &59.3 &{\bf \color{blue} 60.6} &{\bf \color{red} 61.2} \\
			~NfS \cite{NFSdataset} &21.7 &28.0 &35.1 &29.5 &- &- &- &42.2 &- &46.6 &- &50.0 &58.4 &{\bf \color{red} 61.0} & {\bf \color{blue} 60.5}  \\
			\hline
			~Speed (FPS) &{\bf \color{red} 270} &45 &5 &12 &86 &55 &45 &1 &0.3 &8 &{\bf \color{blue} 160} &35 &30 &46 &25 \\
			\hline
		\end{tabular*}
	\end{center}
\end{table*}

\setlength{\tabcolsep}{2pt}
\begin{table}[t]
	\scriptsize
	\begin{center}
		\caption{Comparison with state-of-the-art trackers on the OxUvA \cite{2018longtermBenchmark} dataset in terms of true positive rate (TPR), true negative rate (TNR), and maximum geometric mean (MaxGM). MaxGM is the final evaluation metric.} \label{table:OxUvA}	
		\begin{tabular*}{8.7 cm} {@{\extracolsep{\fill}}lcccccccccccc}
			\hline
			~&MDNet &LCT &TLD &SiamFC+R &MBMD &SPLT &{\bf CARE}\\
			~ &\cite{MDNet} &\cite{LCT} &\cite{TLD} &\cite{SiamFC} &\cite{MBMD} &\cite{SkimmingPerusalTracking} &\\
			\hline
			~MaxGM  &0.343 &0.396  &0.431 &0.454 &0.544 &{\bf \color{blue} 0.622} &{\bf \color{red} 0.749} \\
			~TPR  & 0.472 &0.292  &0.208  &0.427 &{\bf \color{red} 0.609} &0.498 &{\bf \color{red} 0.609} \\
			~TNR  & 0 &0.537  &{\bf \color{blue} 0.895} &0.481 &0.485 &0.776 &{\bf \color{red} 0.922}  \\
			\hline
		\end{tabular*}
	\end{center}
\end{table}

{\flushleft \bf VOT2018 \cite{VOT2018}.} VOT2018 dataset contains 60 challenging videos for short-term tracking evaluation, which will reset the tracker to the ground-truth position when tracking failure occurs. On this benchmark, trackers are evaluated by the Expected Average Overlap (EAO), which considers both accuracy (average overlap over successful frames) and robustness (failure rate).
From Figure \ref{fig:vot}, we can observe that our approach outperforms all the participants on the VOT2018. 
Compared with the recent state-of-the-art approaches, our approach still exhibits satisfactory results.
As shown in Table \ref{table:vot}, our method surpasses the recent regression based methods such as ATOM and DiMP-18 with a relative gain of 2.2\% and 2.0\% in terms of EAO, respectively.
Compared with the cascaded Siamese trackers including SPM \cite{SPM} and C-RPN \cite{CRPN}, our method significantly outperforms them thanks to our online adaptation capability.
Among all the compared trackers, only SiamRPN++ slightly outperforms ours, which adopts a deeper ResNet-50 as the backbone network.

{\flushleft \bf VOT2019 \cite{VOT2019}.} VOT2019 is the recently released challenging benchmark, which replaces 12 easy videos in VOT2018 \cite{VOT2018} by 12 more difficult videos. 
Therefore, the EAO scores of the state-of-the-art trackers such as SiamRPN++ drop sharply.
We compare our approach with the representative approaches in Table~\ref{table:vot19}.
Compared with the SiamRPN++ \cite{siamrpn++} and SiamDW \cite{deeperwiderSiamFC} with deeper ResNet-50, our method with a ResNet-18 obviously surpasses them.
The ATOM \cite{deeperwiderSiamFC} is a top-performing single-stage regression tracker, while ours outperforms it with a relative gain of 6.5\% in terms of EAO.

{\flushleft \bf TrackingNet \cite{2018trackingnet}.} The recent TrackingNet benchmark contains more than 30K videos with more than 14 million dense bounding box annotations. The videos are collected on the YouTube, providing large-scale high-quality data for assessing visual trackers in the wild.
We evaluate our method on the test set of the recently released large-scale TrackingNet dataset, which consists of 511 videos. 
Note that the recent trackers already achieve outstanding AUC scores of more than 70\%, which means the improvement room on this dataset is limited.  
As shown in Table~\ref{table:trackingnet}, the proposed tracker achieves a normalized precision score of 79.0\% and a success score of 71.8\%, which is comparable or superior to previous state-of-the-art trackers such as ATOM and DiMP-18.

{\flushleft \bf Need for Speed \cite{NFSdataset}.} 
NfS dataset contains 100 challenging videos with fast-moving targets, which aims at evaluating the tracking robustness in object fast-moving scenarios.
We evaluate our approach on the 30 FPS version of NfS.
The AUC scores of comparison approaches are shown in Table \ref{table:TC_NFS}. 
Since the search range is limited in SiamRPN++, its performance is relatively unsatisfactory (10.5\% lower than ours in AUC).
The state-of-the-art DiMP-18 and ATOM represent the top performance on this dataset. Our method is comparable with DiMP-18 and outperforms ATOM by 2.1\% AUC.

{\flushleft \bf Temple-Color \cite{TempleColor128}.} Temple-Color benchmark is a challenging dataset consisting of 128 color videos. 
In Table \ref{table:TC_NFS}, we show the AUC score of state-of-the-art trackers on this benchmark.
Compared with the SiamRPN++ with ResNet-50, our method outperforms it by a large margin of 5.0\% AUC score.
The recent single-stage regression tracker ATOM and DiMP yield AUC scores of 59.3\% and 60.6\%, respectively.
The proposed approach also outperforms the recent single-stage regression tracker ATOM and DiMP-18 by 1.9\% and 0.6\% AUC score, respectively. 

\setlength{\tabcolsep}{2pt}
\begin{table*}[t]
	\scriptsize
	\begin{center}
		\caption{Comparison results of state-of-the-art deep trackers with different backbone networks. By adoping a deeper backbone network, our CARE tracker gains further performance improvement and is comparable with the recent state-of-the-art DiMP-50 approach \cite{DiMP}.} \label{table:res50}	
		\vspace{-0.0in}
		\begin{tabular*}{17.8 cm} {@{\extracolsep{\fill}}lcccccccccccc}
			\hline
			~ &Backbone Network &OTB2015 \cite{OTB-2015} &TC128 \cite{TempleColor128} &UAV123 \cite{UAV123} &NfS \cite{NFSdataset} &VOT2018 \cite{VOT2018} &VOT2019 &LaSOT \cite{LaSOT} & TrackingNet \cite{2018trackingnet} &Speed\\
			~ & &AUC score &AUC score &AUC score & AUC score &EAO score &EAO score &AUC score & Success score &FPS\\
			\hline
			~DaSiamRPN \cite{DaSiamRPN} &AlexNet &65.8 &- &58.6 &- &0.326 &- &41.5 &63.8 &{\bf \color{red} 160} \\
			~C-RPN \cite{CRPN} &AlexNet &66.3 &- &- &- &0.289 &- &45.5 &66.9 &32\\
			~SPM \cite{SPM} &AlexNet &68.7 &- &- &- &0.338 &0.275 &- &71.2 &{\bf \color{blue} 110}\\
			\hline 
			~ATOM \cite{ATOM} &ResNet-18 &67.1 & 59.3 & 63.5 &58.4 & 0.401 &0.292 &51.4 &70.3 &30\\
			~DiMP-18 \cite{DiMP} &ResNet-18 &66.2 &60.6 & 63.4 &61.0 &0.402 &- & 53.5 &72.3 &46\\
			~{\bf CARE-18 (Ours)} &ResNet-18 & {\bf \color{blue} 70.5}  &61.2  &{\bf \color{red} 65.4} &60.5 &0.410 &0.323 &54.7 &71.8 &25\\
			\hline
			~SiamRPN++ \cite{siamrpn++}  &ResNet-50 &69.6   &56.2 &61.3 &50.0 &0.414 &0.285 &49.6 &73.3 &35\\
			~DiMP-50 \cite{DiMP}  &ResNet-50 &68.4   &{\bf \color{blue} 61.5}  &64.5 &{\bf \color{blue} 62.0} &{\bf \color{red} 0.440} &{\bf \color{red} 0.379} &{\bf \color{red} 56.9} & {\bf \color{blue} 74.0} &40\\
			~{\bf CARE-50 (Ours)} &ResNet-50 & {\bf \color{red} 71.2}   &{\bf \color{red} 61.7}  &{\bf \color{blue} 64.6} &{\bf \color{red} 62.3} &{\bf \color{blue} 0.427} &{\bf \color{blue} 0.353} &{\bf \color{blue} 56.1} & {\bf \color{red} 74.2} &21\\
			\hline
		\end{tabular*}
	\end{center}
	\vspace{-0.0in}
\end{table*}

{\flushleft \bf UAV20L \cite{UAV123}.} This is a long-term tracking benchmark consisting of 20 long UAV videos with an average length of 2934 frames.
Our second-stage regressor ensures the tracking robustness and helps re-detect the lost target. As a result, our method significantly surpasses previous methods such as ATOM, DiMP-18, and SiamRPN++ (Table \ref{table:TC_NFS}).

{\flushleft \bf OxUvA \cite{2018longtermBenchmark}.} This is a recent large-scale long-term tracking benchmark with 366 videos. The targets in OxUvA undergo frequent partial/full occlusion and out of view. On this dataset, the visual trackers are required to predict the target state (presence or absence) in each frame.
We test our method on the test set of 166 videos. The comparison results are shown in Table~\ref{table:OxUvA}.
We do not add any additional mechanisms (e.g., global search) and merely use the reliability thresholds to predict the target presence/absence.
Note that the recently proposed Skimming-Perusal method (SPLT) \cite{SkimmingPerusalTracking} leads the top performance on this dataset, which is specially designed for long-term tracking with a local-global search.
Without bells and whistles, our approach outperforms SPLT by a relative gain of 20.4\% in terms of MaxGM, showing the importance of online discrimination learning.
The prior motion model (e.g., cosine window) in short-term trackers heavily limits their long-term performance.
Benefited from strong discrimination, our tracker is free of the motion model (e.g., cosine window) and simultaneously handles short-term and long-term scenarios.

\subsection{Performance with a Deeper Backbone Network}
For fair comparison, in this section, we compare our method with state-of-the-art trackers with the same backbone network.
In our approach, we follow ATOM \cite{ATOM} and use a shallow backbone network of ResNet-18 for high efficiency.
By adopting the deeper ResNet-50 \cite{ResNet}, our CARE approach obtains further performance improvements and still maintains a near real-time speed of about 21 FPS on a single Nvidia GTX 1080Ti GPU.
In Table~\ref{table:res50}, we include the recent SiamRPN++ \cite{siamrpn++} and DiMP-50 \cite{DiMP} for comparison, both of which leverage the deep ResNet-50 model.
From the results in Table~\ref{table:res50}, we can observe that our CARE-50 steadily outperforms SiamRPN++ and is comparable with the recent DiMP-50. 
It is worth mentioning that DiMP-50 represents the current state-of-the-art tracker in various tracking benchmarks.

\section{Conclusion} \label{sec:conclusion}
In this paper, we propose a conceptually simple yet effective discrete sampling based ridge regression, which performs as an alternative of the fully-connected layers to discriminate the candidates, but exhibits promising efficiency under a closed-form solution.
Its high flexibility allows the incorporation of hard negative mining as well as our proposed adaptive ridge regression to enhance online discrimination.
We further complement it with the convolutional regression to develop a cascaded framework for robust visual tracking.
The first stage enables a fast and dense search, while the second stage guarantees distractor discrimination.
%
%
The proposed method exhibits outstanding results on several challenging benchmarks with a real-time speed.

\bibliographystyle{IEEEtran}
\bibliography{reference_full}









\end{document}